\documentclass{article}

\PassOptionsToPackage{numbers, compress}{natbib}
\usepackage[final]{neurips_2022}

\usepackage[utf8]{inputenc} % allow utf-8 input
\usepackage[T1]{fontenc}    % use 8-bit T1 fonts
\usepackage[colorlinks = true,
            linkcolor = blue,
            urlcolor  = blue,
            citecolor = blue,
            anchorcolor = blue]{hyperref}
\usepackage{url}            % simple URL typesetting

\usepackage{adjustbox}
\usepackage{booktabs}       % professional-quality tables
\usepackage{amsfonts}       % blackboard math symbols
\usepackage{nicefrac}       % compact symbols for 1/2, etc.
\usepackage{microtype}      % microtypography
\usepackage{bm}
\usepackage{graphicx}
\usepackage{grffile}
\usepackage[hang]{subfigure}
\usepackage{wrapfig}
\usepackage[ruled]{algorithm2e}
\usepackage{threeparttable}
\usepackage{bbding}
\usepackage{caption}
\usepackage{amsfonts}
\usepackage{amsmath}
\usepackage{multirow}
\usepackage[dvipsnames]{xcolor}
\usepackage{xcolor}         % colors
\usepackage{booktabs}       % professional-quality tables
\usepackage{color}
\definecolor{codegreen}{rgb}{0,0.5,0}
\definecolor{codeblue}{rgb}{0,0,0.9}
\definecolor{codeblues}{rgb}{0,0,0.4}
\definecolor{codegray2}{rgb}{0.4,0.4,0.4}
\definecolor{codegray}{rgb}{0.9,0.9,0.9}
\definecolor{codepurple}{rgb}{0.58,0,0.82}
\definecolor{backcolour}{rgb}{0.95,0.95,0.92}
\definecolor{backcolour2}{rgb}{0.9,0.9,0.9}
\definecolor{codered}{rgb}{0.5,0,0}
\definecolor{textcodered}{rgb}{0.05,0.05,0.05}
\definecolor{palegray}{rgb}{0.98,0.98,0.99}
\usepackage{textcomp}

\newcommand{\eg}{\emph{e.g.,}~}
\newcommand{\ie}{\emph{i.e.,}~}
\setcitestyle{square, numbers}
\title{MaskPlace: Fast Chip Placement via Reinforced  Visual Representation Learning
}

\author{%
  Yao Lai \quad Yao Mu \quad  Ping Luo \thanks{Corresponding author is Ping Luo} \\
  Department of Computer Science\\
  The University of Hong Kong\\
  \texttt{\{ylai,ymu,pluo\}@cs.hku.hk} \\
}

\begin{document}

\maketitle

\begin{abstract}
Placement is an essential task in modern chip design, aiming at placing millions of circuit modules on a 2D chip canvas. Unlike the human-centric solution, which requires months of intense effort by hardware engineers to produce a layout to minimize delay and power consumption, deep reinforcement learning has become an emerging autonomous tool.
However, the learning-centric method is still in its early stage, impeded by a massive design space of size ten to the order of a few thousand. 
This work presents MaskPlace to automatically generate a valid chip layout design within a few hours, whose performance can be superior or comparable to recent advanced approaches. It has several appealing benefits that prior arts do not have.
Firstly, MaskPlace recasts placement as a problem of learning pixel-level visual representation to comprehensively describe millions of modules on a chip,  enabling placement in a high-resolution canvas and a large action space. It outperforms recent methods that represent a chip as a hypergraph.
Secondly, it enables training the policy network by an intuitive reward function with dense reward, rather than a complicated reward function with sparse reward from previous methods.
Thirdly, extensive experiments on many public benchmarks show that MaskPlace outperforms existing RL approaches in all key performance metrics, including wirelength, congestion, and density.
For example, it achieves 60\%-90\% wirelength reduction and guarantees zero overlaps.
We believe MaskPlace can improve AI-assisted chip layout design.
The deliverables are released at \href{https://laiyao1.github.io/maskplace/}{laiyao1.github.io/maskplace}.
\end{abstract}

\section{Introduction}

\begin{figure}[t] 
% \captionsetup[subfigure]{justification=centering}
	\centering 

	\subfigure[DREAMPlace \cite{lin2020dreamplace}]{
% 		\label{level.sub.3}
		\includegraphics[width=0.23\linewidth]{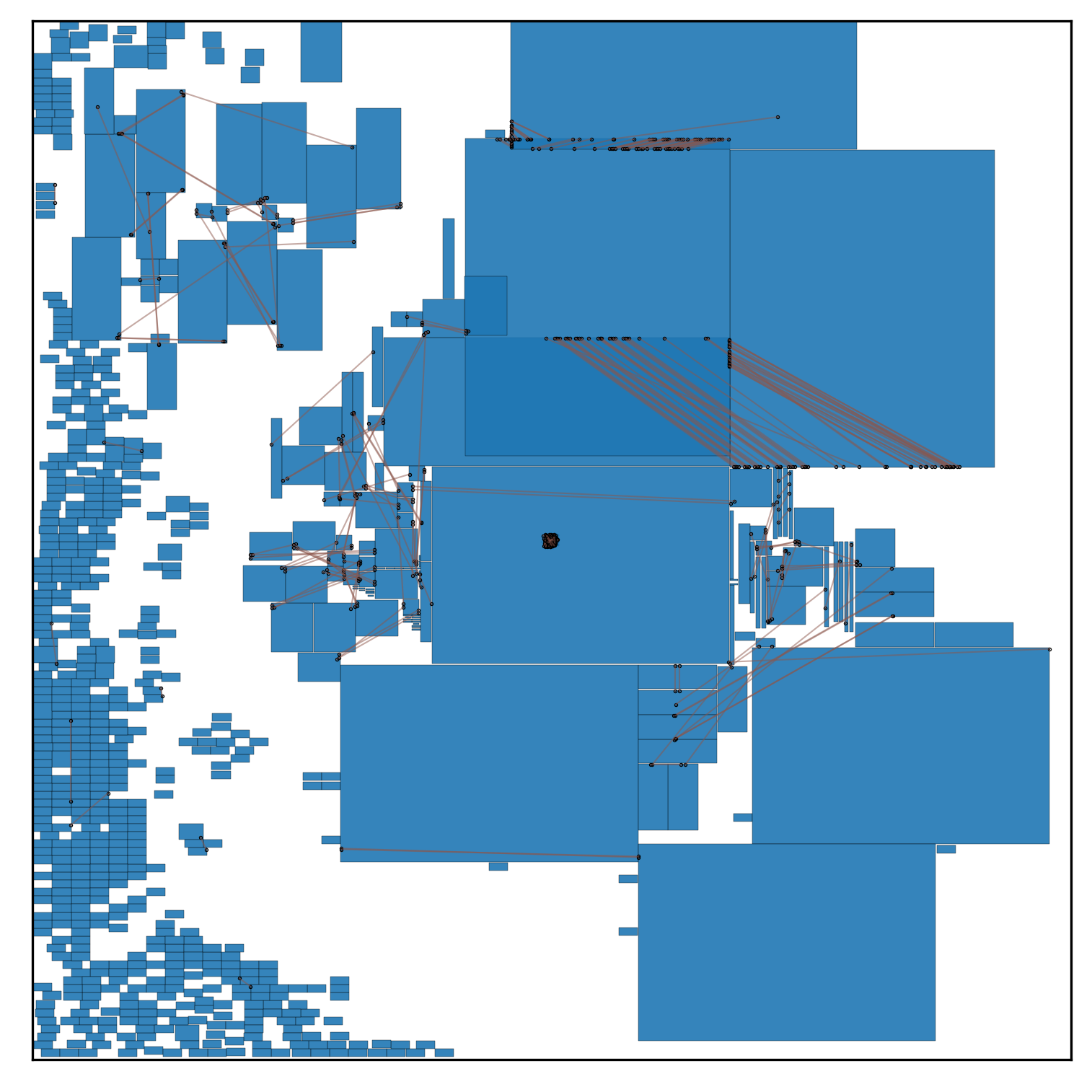}}
% 	\quad
	\subfigure[Graph Placement \cite{mirhoseini2021graph} ]{
% 		\label{level.sub.4}
		\includegraphics[width=0.23\linewidth]{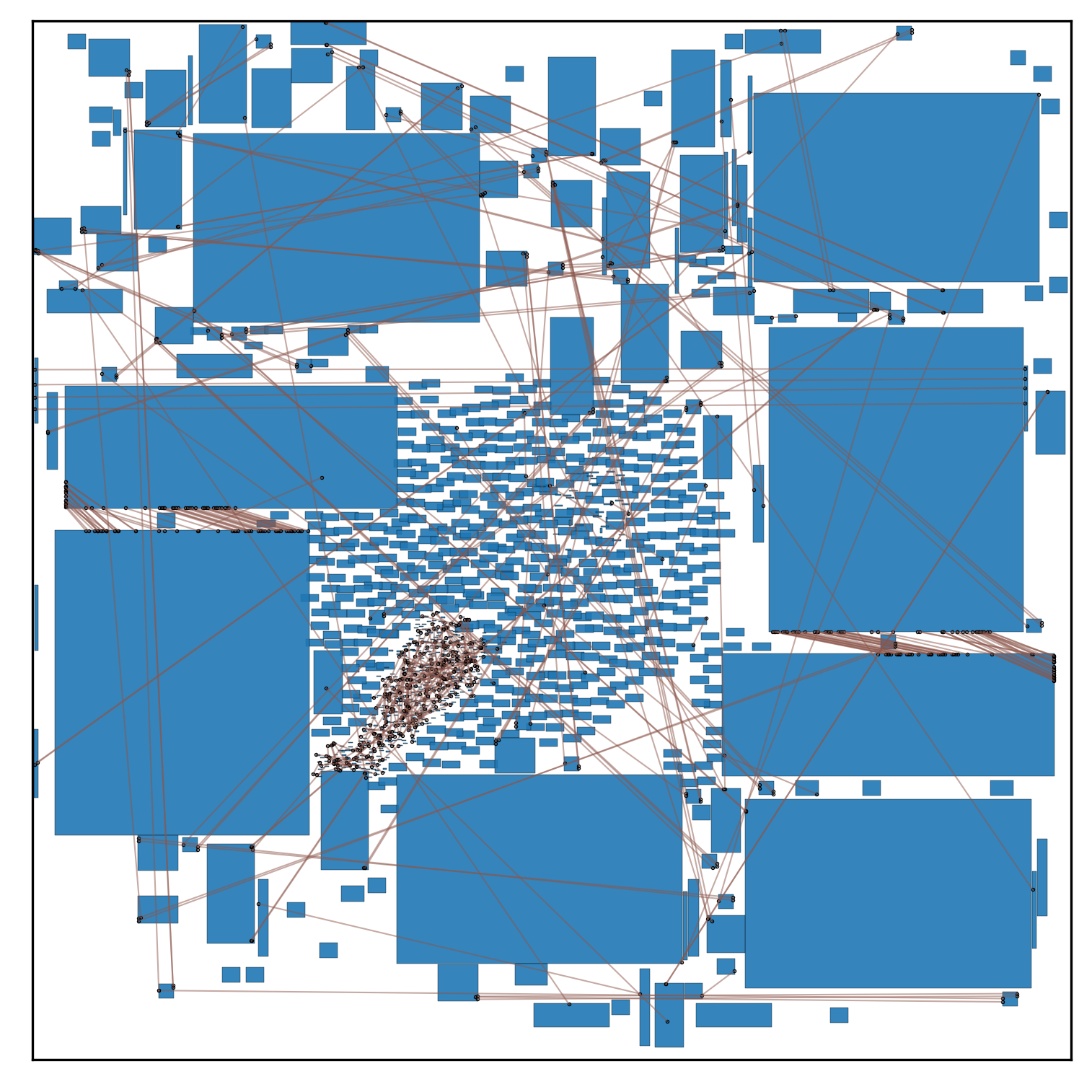}}
	\subfigure[DeepPR \cite{cheng2021joint} ]{
% 		\label{level.sub.4}
		\includegraphics[width=0.23\linewidth]{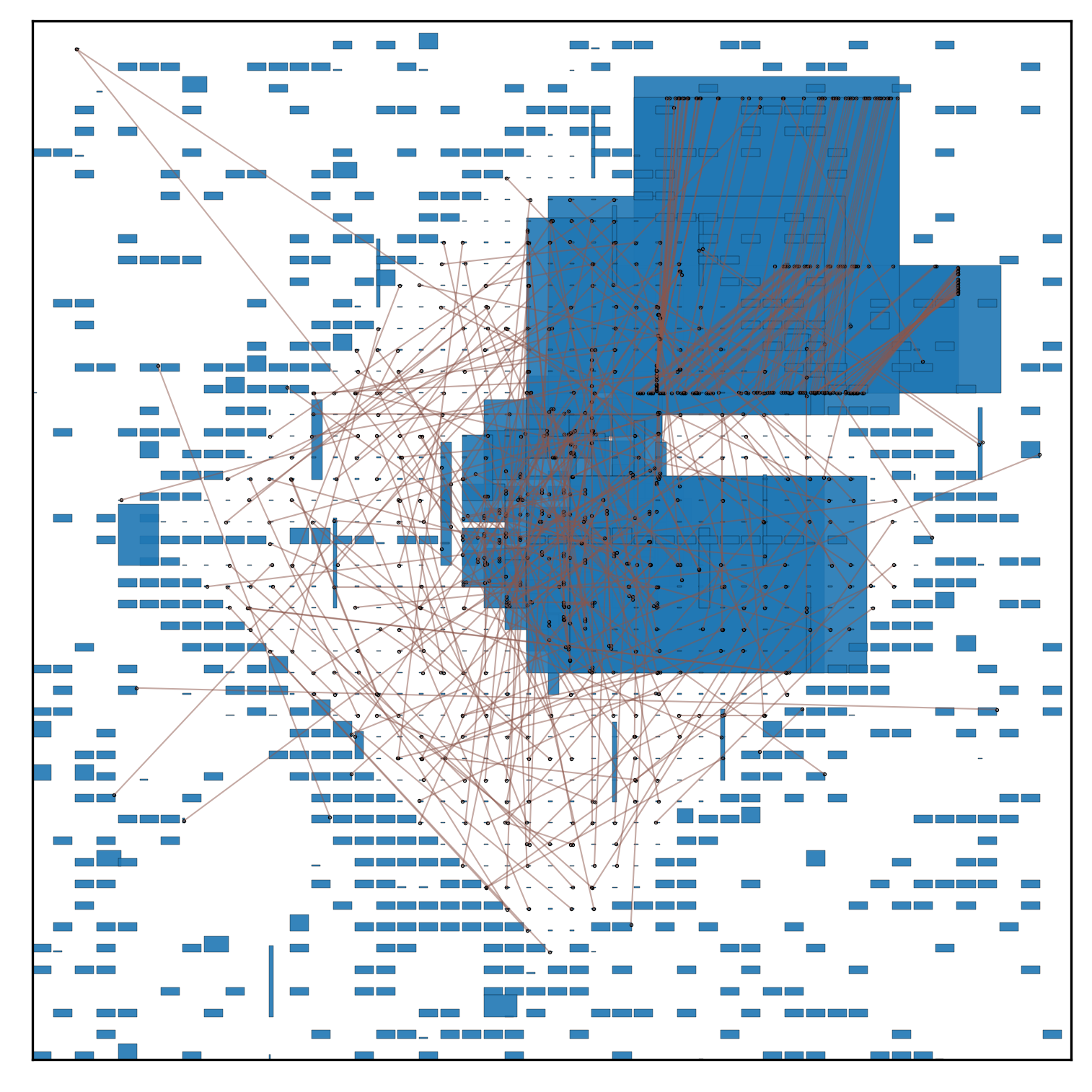}}
	\subfigure[MaskPlace (ours)]{
% 		\label{level.sub.4}
		\includegraphics[width=0.23\linewidth]{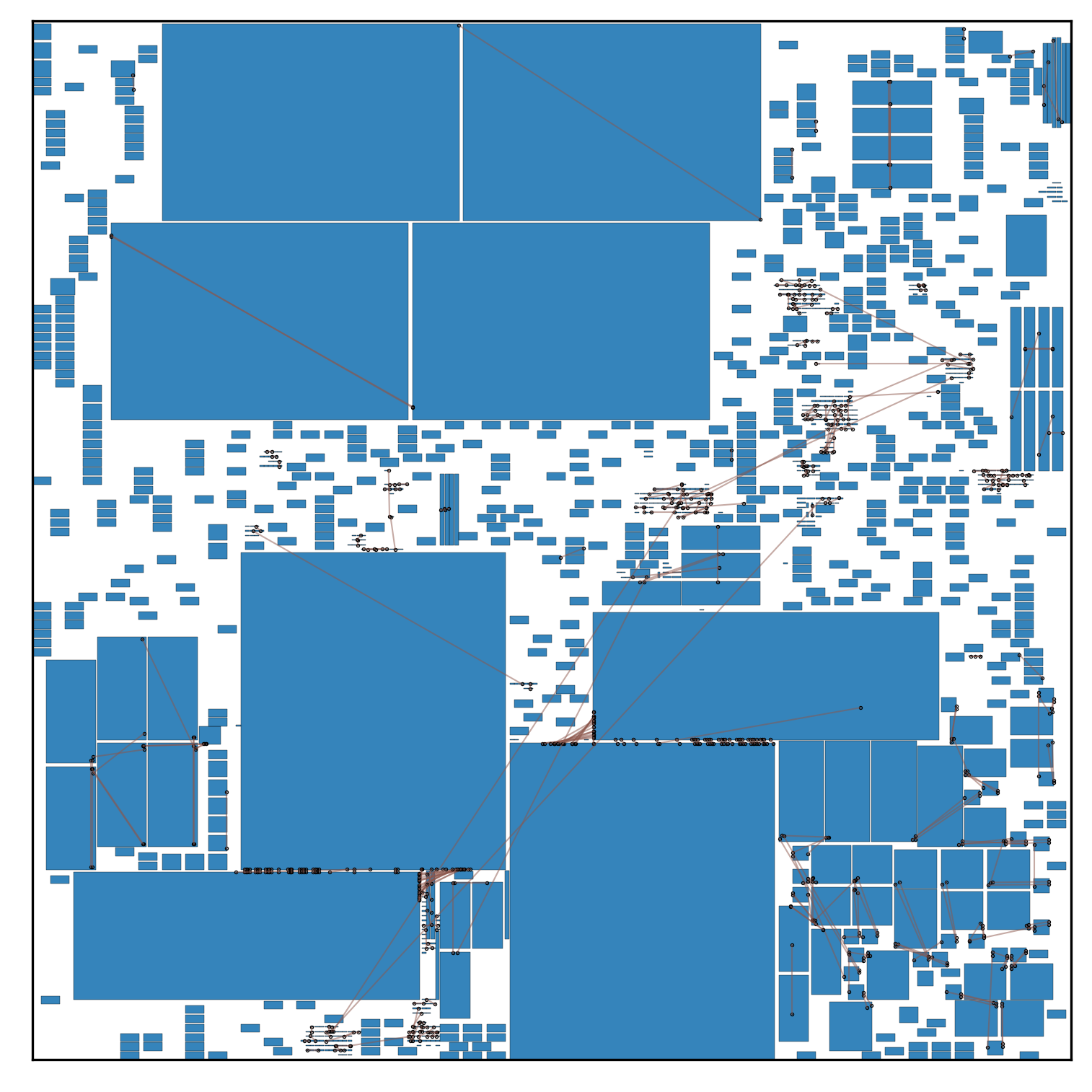}}
\caption{\small{ \textbf{Visualizing different placements of a circuit benchmark} \textit{bigblue3}, where the modules are visualized by \textcolor{MidnightBlue}{blue} rectangles and the wires are shown in \textcolor{brown}{brown} lines to connect massive pins on modules. For clarity, we only show 1\% wires. The proposed MaskPlace is compared with three representative approaches, including \textbf{(a)} DREAMPlace \cite{lin2020dreamplace} (HPWL $=1.04\times 10^7$, WL $=1.08\times 10^7$, OL $=8.06\%$), \textbf{(b)} Graph Placement \cite{mirhoseini2021graph} (HPWL $=3.45\times 10^7$, WL $=3.73\times 10^7$, OL $=0.80\%$), \textbf{(c)} DeepPR \cite{mirhoseini2021graph} (HPWL $=4.39\times 10^7$, WL $=5.18\times 10^7$, OL $=85.23\%$), and  \textbf{\textcolor{red}{(d) MaskPlace}} (HPWL $=\underline{0.83\times 10^7}$, WL $=\underline{0.88\times 10^7}$, OL $=\underline{0\%}$), where HPWL, WL, and OL represent half-perimeter wirelength\protect\footnotemark, wirelength, and overlap area ratio, respectively. All the metric values are smaller the better. The best performances are underlined in (d). We  see that MaskPlace surpasses the recent popular placement approaches  in all key metrics, and it can satisfy the $0\%$ hard density constraint. \textbf{Better zoom in 400\%.}} }
% \small\textsuperscript{a=} The footnote-like comment under the caption
\label{fig:placefig}
\end{figure}

% , such as computers and mobile phones
% Chips are used in a wide range of devices nowadays. 
% Chips are used in a wide variety of devices.
The scalability and efficiency are two significant factors of autonomous chip layout design. Placement is one of the most challenging and time-consuming problems in the design flow, aiming to determine the locations of millions of circuit modules on a 2D chip canvas represented by a two-dimensional grid. 
A netlist can describe these modules, that is, a large-scale hypergraph consisting of massive macros (functional blocks such as memory) and standard cells (logic gates), where each macro and each standard cell can contain several or even hundreds of pins connected by wires, as shown in Fig.\ref{fig:placefig}.

Placing a large number of circuit modules onto the chip canvas is challenging because many performance metrics such as power consumption, timing, area, and wirelength should be minimized while satisfying some hard constraints such as placement density and routing congestion.
For example, the wirelength (the length of wires that connect all modules) determines the delay and the power consumption of a chip \cite{wang2009electronic}. Shorter wires often indicate less delay and less power consumption \cite{rabaey2002digital}. However, wirelength cannot be reduced by overlapping modules because the module density is a hard constraint to ensure that a valid and manufacturable chip layout has non-overlapping modules.
More examples of the performance metrics are given in Fig.\ref{metrics} and Fig.\ref{modulenetpin} in Appendix.
As pointed out in \cite{mirhoseini2021graph}, the design space of placement is larger than $10^{2,500}$ when there are just $1,000$ circuit modules, whereas neural architecture search (NAS) typically has a  space of $10^{30}$ and the Go game has a state space of $10^{360}$.

%modules cannot overlap with each other according to the requirements of chip manufacturing.

%The development of chip design technology can improve chip performance. From the chip design perspective, a chip contains millions of modules. Wires connect these modules, and the wirelength determines the delay and power of the chip \cite{wang2009electronic}. Therefore, how to choose positions to place is a critical issue. In the placement, we want to minimize total wirelength because shorter wires mean less delay and less power-consuming capacitive loads \cite{rabaey2002digital}. Moreover, modules cannot overlap with each other according to the requirements of chip manufacturing. The search space for placement is exponential, and it is more than $10^{2,500}$ when there are 1,000 modules  \cite{mirhoseini2021graph}, so it is still challenging.
% on the chip
% , including relatively large macros (memories, IO interfaces) and small standard cells (logical gates)
% the size of the macros cannot be ignored, and they
% according to the logical design
% in the chip design
% within a congestion threshold

Methods of chip placement can be generally divided into two categories, classic optimization-based approaches \cite{roy2006min, khatkhate2004recursive, chen2008ntuplace3, lu2014eplace, cheng2018replace, lin2020dreamplace, yang2000dragon2000, vashisht2020placement, viswanathan2007rql, viswanathan2007fastplace, kim2012maple, kim2012complx, brenner2015bonnplace, lin2013polar, spindler2008kraftwerk2, chan2006mpl6, kahng2005aplace, gu2020dreamplace} and learning-based approaches \cite{mirhoseini2021graph, cheng2021joint, jiang2021delving}.
In the first category, hardware scientists often formulate placement as an optimization problem and relax the hard constraints.  For example, let a pair of vectors $(\bm{x},\bm{y})$ denote the ($x,y$)-coordinate value of all circuit modules on a 2D canvas, the objective function of  placement  can be formulated as minimizing $\mathrm{WL}(\bm{x}, \bm{y})$, subject to $\mathrm{D}(\bm{x}, \bm{y})\le \alpha$, where $\mathrm{WL}(\cdot,\cdot)$ and $\mathrm{D}(\cdot,\cdot)$ are the estimation functions of wirelength and density respectively, and $\mathrm{D}(\bm{x}, \bm{y})\le \alpha$ is a hard constraint with a very small density value $\alpha$, which ensures that all modules do not overlap.
For instance, DREAMPlace \cite{lin2020dreamplace} is a recent advanced method that minimizes $\mathrm{WL}(\bm{x}, \bm{y})+\lambda \mathrm{D}(\bm{x}, \bm{y})$, which relaxes the hard density constraint. However, it cannot directly produce a valid and manufacturable layout because the non-overlapping constraint is not satisfied after relaxation. 
These approaches often need a post-processing step, such as manual refinement and  legalization (LG), to remove the overlapping in placement, resulting in two issues,  
(1) the wirelength may increase substantially after LG, and (2) no feasible solution can be found if the available chip area is insufficient before post-processing.

\footnotetext{HPWL (Half Perimeter Wire Length) is a common approximation metric of the wirelength and can be computed much more efficiently than wirelength.}

% \begin{wrapfigure}[20]{r}{0.45\textwidth}
%   \begin{center}
%     \includegraphics[width=0.44\textwidth]{realvisual_new.pdf}\\
%   \end{center}
%   \caption{Placement and Mask Visualization}
%   \label{visualzation}
% \end{wrapfigure}

In the second category, reinforcement learning (RL) is employed 
to solve placement as a sequential decision-making problem, placing each circuit module at a time.  
Although the learning-based approaches are still in their early stage, they can produce promising results to automate the chip design flow end-to-end significantly without human effort.
For instance, Graph Placement \cite{mirhoseini2021graph} and DeepPR \cite{cheng2021joint} represent a netlist as a hypergraph, denoted as $G=(V, E)$, where $V$ represents a set of nodes, and each node is a module, and $E$ is a set of edges, which are the wires connecting all modules. They train RL agents to place one module at a time by maximizing the metric values as rewards.
However, the hypergraph is not scalable to comprehensively encode information of a netlist. For example, the relative positions (offsets) of pins are discarded in \cite{mirhoseini2021graph, cheng2021joint}. The wirelength estimation is inaccurate without the pin information, but encoding this rich information would make the hypergraph too complicated because each module can have hundreds of pins.
Furthermore, placement on a large hypergraph requires heavy computations. \citet{mirhoseini2021graph} reduced computations by placing 15\% of the modules using reinforcement learning (the remaining modules are placed by classic method), and \citet{cheng2021joint} decreased the size (resolution) of module and chip canvas as shown in Table \ref{table:method compare}. Both of them sacrificed their placement performance.

To address the issues of prior arts,
we propose a novel RL method, named MaskPlace, which can automatically generate a high-quality and valid layout (non-overlapping modules)  within a few hours, unlike previous methods that need manual refinement to modify invalid placement, which may wait up to 72 hours for commercial electronic design automation (EDA) tools to evaluate the placement.
MaskPlace casts placement as a problem of pixel-level visual representation learning for circuit modules using convolutional neural networks. This representation can comprehensively capture the configurations of thousands of pins, enabling fast placement in a full action space on a large canvas size \eg 224$\times$224.
As shown in Fig.\ref{fig:placefig} and Table \ref{table:method compare}, MaskPlace has many attractive benefits that existing works do not have. MaskPlace is mainly for macro placement due to the problem size.

This paper has three main \textbf{contributions}. Firstly, we recast chip placement as a problem of learning visual representation to describe millions of circuit modules on a chip comprehensively. It opens up a new perspective for AI-assist chip placement.
Secondly, we carefully design a new policy network that can capture and aggregate both the global and subtle information on a chip canvas, maximizing the reward of wirelength and ensuring non-overlapping placement efficiently.
Thirdly, extensive experiments demonstrate that MaskPlace outperforms recent advanced methods on 24 public chip benchmarks. For example,  MaskPlace can always produce a layout with 0\% overlap while reducing wirelength up to 5$\times$ and 9$\times$ compared to Graph Placement \cite{mirhoseini2021graph} and DeepPR \cite{cheng2021joint} respectively. 
% Secondly, we carefully design a new policy network that can capture and aggregate both the global and subtle information on a chip canvas, enabling us to maximize the rewards of wirelength and ensure non-overlapping placement efficiently.

%To make the chip more practical and have better metrics, we propose a new RL method MaskPlace. It is learned by various visual representations, including position mask, wirelength mask, and view mask as Fig. \ref{visualization}. These masks provide the information about position, wirelength and overall status for placement. They can prevent overlaps and introduce the pin offset information. 
% , which are introduced with prior knowledge
% The addition of prior knowledge 
%They allow our model to find a satisfactory placement solution efficiently in a complete search space. Also, to prevent position information from disappearing during the convolution, we use the 1x1 conv layers to extract local features on masks. A detailed comparison between the methods can be seen in Tab. \ref{method compare}. The placement results from one circuit are shown in Fig. \ref{fig:placefig}.

\begin{table}[!t]
\centering
\scriptsize
\caption{ \small{\textbf{Comparisons} of representative placement methods in different aspects, including method types  (``Family''), canvas size  (``Resolution''), state space, ``0\% overlap'' (if the method can produce a layout without overlapping placement), training/inference speed (``Efficiency''), and the performance metrics to be optimized. We see that MaskPlace can outperform recent advanced methods by performing placement on a full canvas size of 224$\times$224 (much larger than prior works) and producing a valid placement with 0\% overlap (which cannot be achieved by previous methods). MaskPlace can also be trained and tested efficiently.}}\label{table:method compare}
\begin{threeparttable}
\begin{tabular}{cccccccccc}\\
\toprule  
% Method  & DREAMPlace  &  Graph Placement & DeepPR & MaskPlace\\\midrule
% Family   & Nonlinear & RL + Nonlinear & RL & RL \\ \midrule
% Resolution &  Continuous & $128^2$ & $32 ^2$& $224^2$\\ \midrule
% Action Space &  - & ${(128^2)}^{\alpha V}, \alpha \approx 0.1$ & ${(32^2)}^V$ & ${(224^2)}^V$\\ \midrule

% No overlap & \XSolidBrush (need LG) & \XSolidBrush (only hard macros) & \XSolidBrush & \CheckmarkBold \\ \midrule
% Real Size  &   \CheckmarkBold & \CheckmarkBold & \XSolidBrush &\CheckmarkBold\\\midrule
% Train Efficient & - & Medium & High & High\\ \midrule
% Infer Efficient &  High & Medium & Medium & High  \\ \midrule
% Metrics & H,D &  H,C,D & H,C & H,C,D\\ 
~ & Family & Resolution & State Space & $0\%$ Overlap %& Real Size 
& Reward & Efficiency  & Metrics\\
\midrule
DREAMPlace \cite{lin2020dreamplace}& Nonlinear & Continuous & - & \XSolidBrush \tnote{~\textcolor{blue}{1}} %& \CheckmarkBold 
& - & - /High & H, D \tnote{~\textcolor{blue}{2}}\\
Graph Placement \cite{mirhoseini2021graph}& RL+Nonlinear & $128^2$ & ${(128^2)}^{\alpha V}$\tnote{~\textcolor{blue}{3}} & \XSolidBrush 
%& \CheckmarkBold 
& Sparse & Med./Med. & H, C, D\\
DeepPR \cite{cheng2021joint} & RL & $32 ^2$& ${(32^2)}^V$ & \XSolidBrush 
%& \XSolidBrush 
& Dense & High/Med. & H, C\\
MaskPlace (ours) & RL & $224^2$ & ${(224^2)}^V$ & \CheckmarkBold 
%& \CheckmarkBold 
& Dense & High/High & H, C, D\\
\bottomrule
\end{tabular}
\begin{tablenotes}\tiny
\item[1] DreamPlace needs a post-processing step, such as  legalization (LG) that may fail.
\item[2] H = HPWL, C = Congestion, D = Density.
\item[3] $V$ is the number of circuit modules and $\alpha \approx 15\%$ in Graph Placement.
%and only $\alpha V$ number of modules (hard macros) do not overlap each other.
\end{tablenotes}
\end{threeparttable}
\end{table}

\section{Preliminary and Notation}

\begin{figure}[t]
\centering
% \pluo{rewrite caption make it self-contain, expalin and discuss more.}
\includegraphics[width=0.95\textwidth]{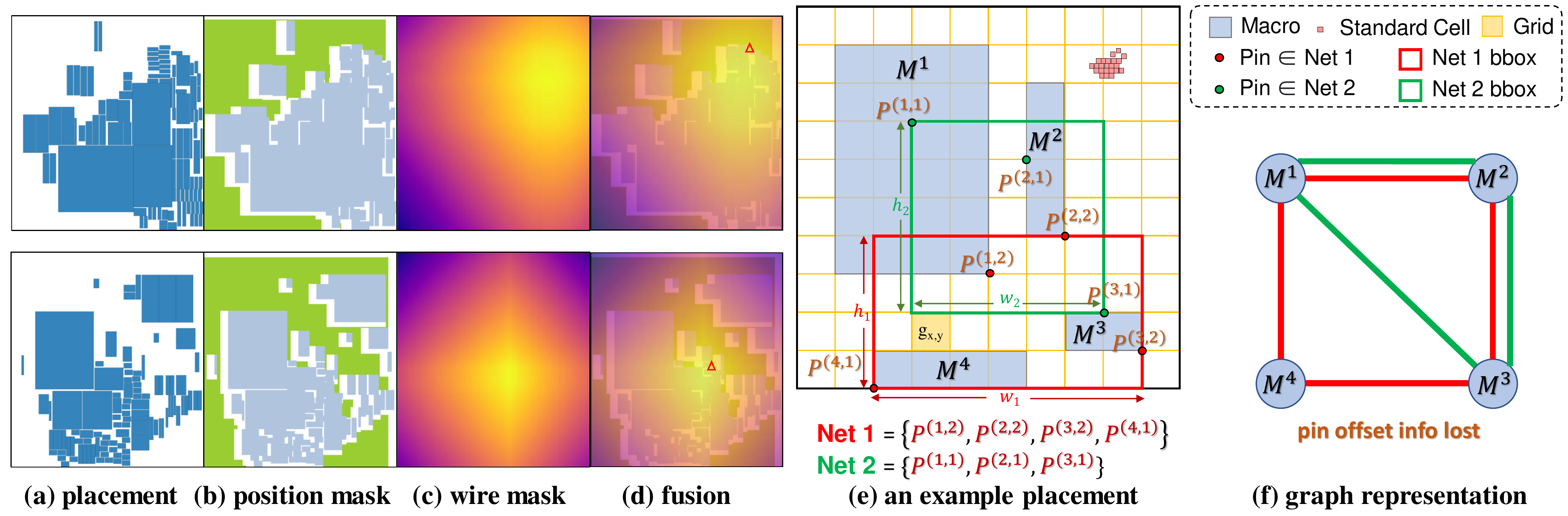}
\caption{\small{\textbf{Mask Visualization, placement example, and hypergraph representation in prior work.}  We visualize different masks in MaskPlace (a-d) and illustrate an example of placement in (e). In the position mask (b), the \textcolor{LimeGreen}{green} color means feasible positions to place while the \textcolor{CadetBlue}{gray} color represents the placed modules. In the wire mask (c), lighter color indicates shorter wirelength if a module is placed at a specific position. The fusion mask in (d) is an example of the output after the mask fusion model using $1\times1$ convolutions, where the \textbf{\textcolor{red}{$\triangle$}} denotes the position with a high probability to place at (\ie no overlap and shorter wirelength). (f) is the result when converting the circuit in (e) into a hypergraph in prior works, where the critical information of pin locations is lost.} 
%Fusion is the information exchange between position mask and wirelength mask. \textcolor{red}{$\triangle$} is the position with high probability to place (no overlap and shorter wirelength).
}
% \vspace{-8pt}
\label{placementgraph}
\end{figure}

The placement quality can be measured by the HPWL (half perimeter wirelength), which estimates the wirelength with marginal computational cost \cite{chen2006high}.  Intuitively, Fig.\ref{placementgraph}(e) illustrates a 2D chip canvas. Let $M^i$ and $P^{(i,j)}$ denote the $i$-th module and its $j$-th pin, respectively. A net contains a set of pins connecting modules by wires. For example, ``Net 1'' (in red) connects all four modules (\ie $M^1,M^2,M^3,M^4$)  using wires through pins $P^{(1,2)}$, $P^{(2,2)}$, $P^{(3,2)}$, and $P^{(4,1)}$, while  ``Net 2'' (in green) connects three modules (\ie $M^1,M^2,M^3$)  using wires through pins $P^{(1,1)}$, $P^{(2,1)}$, and $P^{(3,1)}$.
HPWL estimates the wirelength by summing up the half perimeters of bounding boxes of all the nets, as shown by the red and green boxes in Fig.\ref{placementgraph}(e). Intuitively, the half perimeter of a net bounding box equals the sum of its height and width.
For example, HPWL in Fig.\ref{placementgraph}(e) is $h_1+w_1+h_2+w_2$.
%there are two nets (``Net 1'' and ``Net 2''), where each net connects a set of pins using wires.  
%
%Intuitively, 
%where the bounding box is the minimal rectangle including all pins belonging to this net.
%

% Given a netlist containing a set of nets, minimizing the wirelength can be treated  as minimizing HPWL  by placing modules to the optimal  positions on a 2D chip canvas, that is, $\min $

% $\sum_{\forall\mathrm{net} \in \mathrm{netlist}} \big( \max_{P^{(i,j)} \in \mathrm{net}} P^{(i,j)}_x-\min_{P^{(i,j)} \in \mathrm{net}} P^{(i,j)}_x + \max_{P^{(i,j)} \in \mathrm{net}} P^{(i,j)}_y-\min_{P^{(i,j)} \in \mathrm{net}} P^{(i,j)}_y \big)$,
% where  $P_x$ and $P_y$ represent the $(x,y)$-coordinate value of  a pin, respectively.

% To achieve a valid and manufacturable chip layout, we need to satisfy two hard \textbf{constraints}, (1) the wire congestion should be lower than a desired small threshold to reduce chip cost (\ie $Congestion(M_x, M_y, M_w, M_h) \le C_{th}$, where $M_x, M_y, M_w, M_h$ represent the position, width, and height of modules), and (2) the density should be minimized to achieve non-overlapping placement (\ie $Overlap(M_x, M_y, M_w, M_h) =0$).

Given a netlist containing a set of nets, minimizing the wirelength can be treated as minimizing HPWL by placing modules to the optimal positions on a 2D chip canvas. To achieve a valid and manufacturable chip layout, we need to satisfy two hard constraints: (1) \textit{congestion constraint}: the wire congestion should be lower than a desired small threshold to reduce chip cost, and (2) \textit{overlap constraint}: the density should be minimized to achieve non-overlapping placement.
\begin{equation}
\footnotesize
\begin{split}
\min \quad &\sum_{\forall\mathrm{net} \in \mathrm{netlist}} \big( \max_{P^{(i,j)} \in \mathrm{net}} P^{(i,j)}_x-\min_{P^{(i,j)} \in \mathrm{net}} P^{(i,j)}_x + \max_{P^{(i,j)} \in \mathrm{net}} P^{(i,j)}_y-\min_{P^{(i,j)} \in \mathrm{net}} P^{(i,j)}_y \big)   \\
\text{s.t.}\quad & \mathrm{Congestion} (M_x, M_y, M_w, M_h)\le C_{\mathrm{th}} \quad \mathrm{and} \quad  \mathrm{Overlap} (M_x, M_y, M_w, M_h) =0,\\
\end{split}
\end{equation}
where  $P_x$ and $P_y$ represent the $(x,y)$-coordinate value of  a pin respectively, $\mathrm{Congestion}(\cdot)$ is the congestion function, $C_{\mathrm{th}}$ is a desired threshold,  $\mathrm{Overlap}(\cdot)$ is the overlap function, and $M_x, M_y, M_w, M_h$ represent the position, width, and height of modules respectively.
Firstly,  lower congestion often indicates shorter wirelength, which is crucial to reduce chip cost because the wire resources are limited on a real chip. 
Inspired by prior arts \cite{mirhoseini2021graph, cheng2021joint}, we employ the RUDY estimator \cite{spindler2007fast} to estimate wire congestion. Details of RUDY can be found in the Appendix \ref{detailmetric}. 
Secondly, the placement density calculates the overlapping region between every pair of circuit modules. It is time-consuming since its computational complexity is $\mathcal{O}(V^2)$ where $V$ is the number of modules \cite{wang2009electronic}.
The proposed approach can ensure non-overlapping placement to avoid calculating this density metric explicitly in training, thus reducing computations while producing a valid layout.

%Routing is a process to determine the precise wire paths for nets. 
% So, if congestion is too high, wires may not be routed optimally, and wirelength will increase. 
% There are many ways to estimate congestion, one is to compute a routing result \cite{mirhoseini2021graph}, but it is very computationally intensive. 
%We use RUDY \cite{spindler2007fast} as the estimation of congestion, which is a common way to evaluate.
% , which is same as DeepPR\cite{cheng2021joint} used. 
%In RUDY, each grid needs to accumulate the inverse of the height and width $(1/h+1/w)$ of all the net bounding box covering itself, and take out the maximum value (or the average of the first k maximums) of all grids.
%\paragraph{Density}
% with the number of modules $V$ 

\section{Our Approach}

\begin{figure}[!t]
\centering
\includegraphics[width=0.95\textwidth]{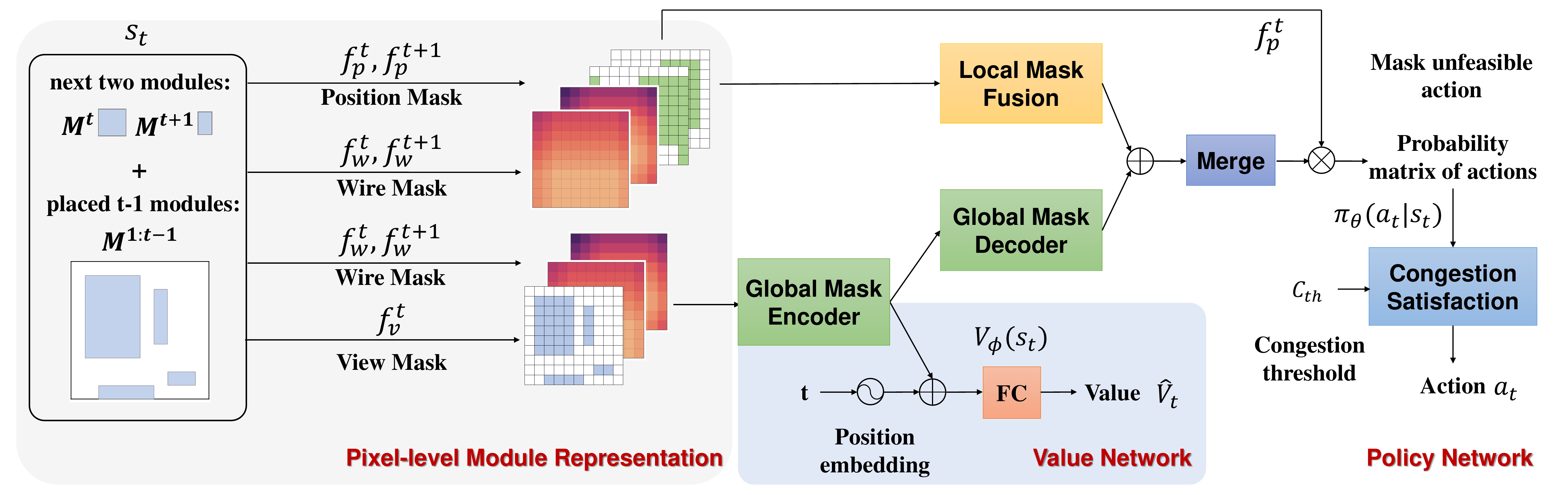}
% \pluo{rewrite caption make it self-contain, expalin and discuss more.}
\caption{\small{ \textbf{Overview of MaskPlace,} which contains three main parts: a pixel mask generation model, a policy network, and a value network. The pixel mask generation model converts the current placement state into pixel-level masks. The policy and value networks convert these masks to actions and values based on global and local features. The congestion satisfaction block is to satisfy the congestion constraint and give the final action.} }
\label{architecture}
\vspace{-8pt}
\end{figure}

\textbf{Model Architecture Overview.}
Chip placement can be formulated as a Markov Decision Process (MDP) \cite{sutton2018reinforcement} by placing each module at a time.
Fig.\ref{architecture} illustrates the overall architecture of MaskPlace, which trains a policy $\pi_{\theta}(a_t|s_t)$ represented by a convolutional encoder-decoder network with parameter set $\theta$, and a value function $V_{\phi}(s_t)$ represented by an embedding model with parameter set $\phi$. 
The policy network receives previous observations and actions as input $s_t$ and selects an action $a_t$ as output. Specifically, $s_t$ is a set of pixel-level feature maps that comprehensively capture the net and pin configurations in $M^{1:t-1}$, $M^t$, and $M^{t+1}$, where $M^{1:t-1}$ denotes the modules that have been placed in the previous time steps from $1$ to $t-1$, while $M^t$ and $M^{t+1}$ denote  the modules to be placed at the current step $t$ and the next step $t+1$, respectively. Intuitively, MaskPlace looks one step forward to achieve better placement.

\begin{wrapfigure}[15]{r}{0.32\textwidth}
%   \vspace{-0.4cm}
  \begin{center}
    \includegraphics[width=0.3\textwidth]{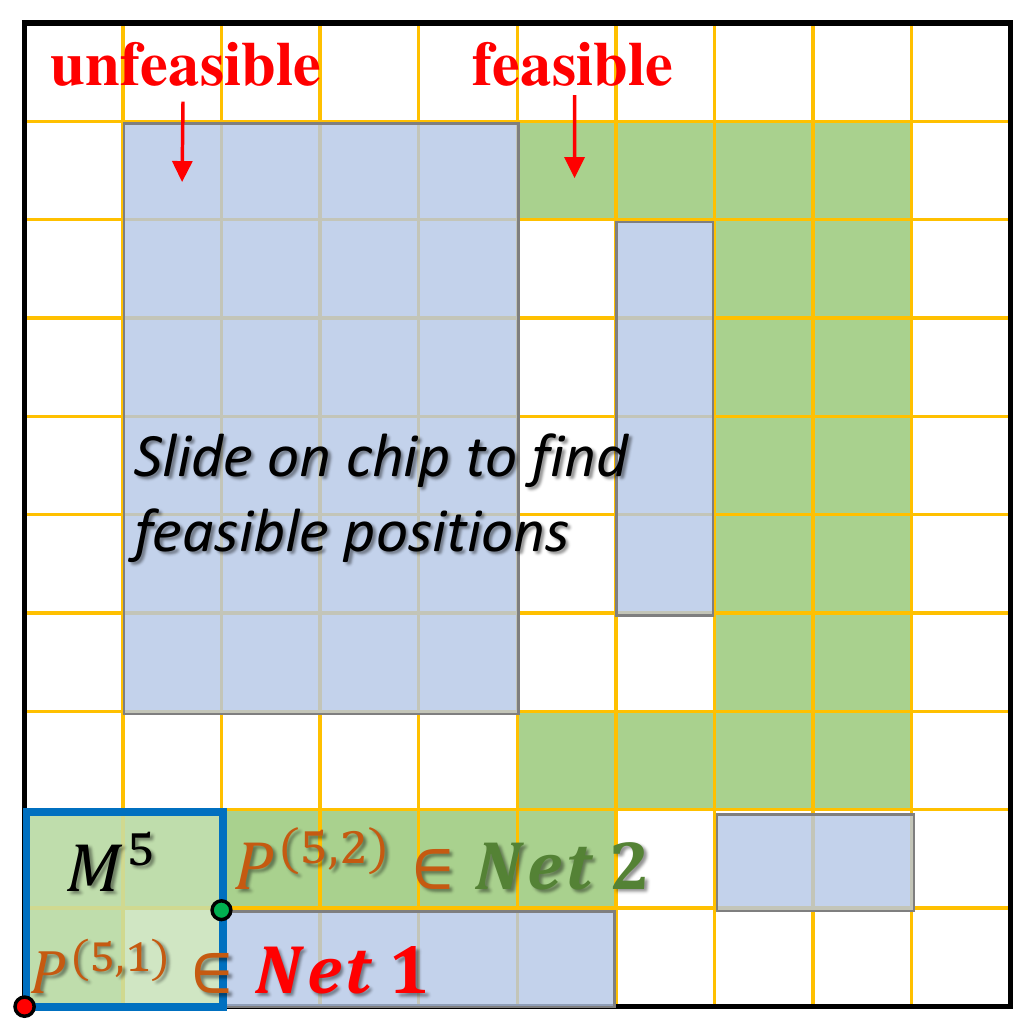}
  \end{center}
  \caption{\small{Position Mask Example.}}
  \label{posmask}
\end{wrapfigure}

Although prior arts \cite{mirhoseini2021graph, cheng2021joint} represented a netlist as a hypergraph as shown in Fig.\ref{placementgraph}(f) where each node is a module, and each edge is a wire between two modules, they lost the information of pin offsets for each module.
Unlike previous works, MaskPlace can fully represent massive 
net and pin configurations using three types of pixel-level feature maps, as shown in Fig.\ref{placementgraph}(a-d), including position mask, wire mask, and view mask, as discussed below. Different masks are fused by convolutions to learn the state representation.
%if we determine the placement order and then place them one by one.

\textbf{Position Mask.} The position mask, denoted by $f_p \in \{0, 1\} ^{224\times 224}$, is a binary matrix of a canvas grid with size $224\times 224$ as shown in Fig.\ref{posmask}, where value ``$1$''  means a feasible position to place a module. The purpose of the position mask is to guarantee no overlaps between modules (\ie satisfy the overlap constraint) and to learn the relationship between placement and wirelength. 
Specifically, we slide a module $M^t$ (for example, $t=5$) on the entire chip canvas. The trajectory of the feasible positions (in green)
can be labeled with ``1''. Intuitively, we can check each position for each module using the cumulative sum array \cite{guo2021ultrafast}. This naive approach has the computational complexity of $\mathcal{O}(N^2)$ when a 2D canvas grid is divided into $N\times N$ cells.
However, this simple approach is not efficient when $N$ is large. Therefore, since all modules are rectangles, we design an efficient generation algorithm,  
which iterates through all placed modules (in blue) and excludes positions that will cause overlap.
In this case, all remaining positions are available for placement.
The new algorithm is summarized in Appendix \ref{algorithm}, which costs $\mathcal{O}(V)$ for each module, where $V$ is the number of modules.
%and $O(V^2)$ for one placement when matrix assignment has constant cost.
%formed by the marker (at the left-bottom grid of the module) 
%is the side length of the canvas.
% It finds all infeasible positions due to overlap for each placed module and takes the union set of all infeasible positions. 

\begin{wrapfigure}[15]{r}{0.32\textwidth}
    % \vspace{-0.6cm}
  \begin{center}
    \includegraphics[width=0.32\textwidth]{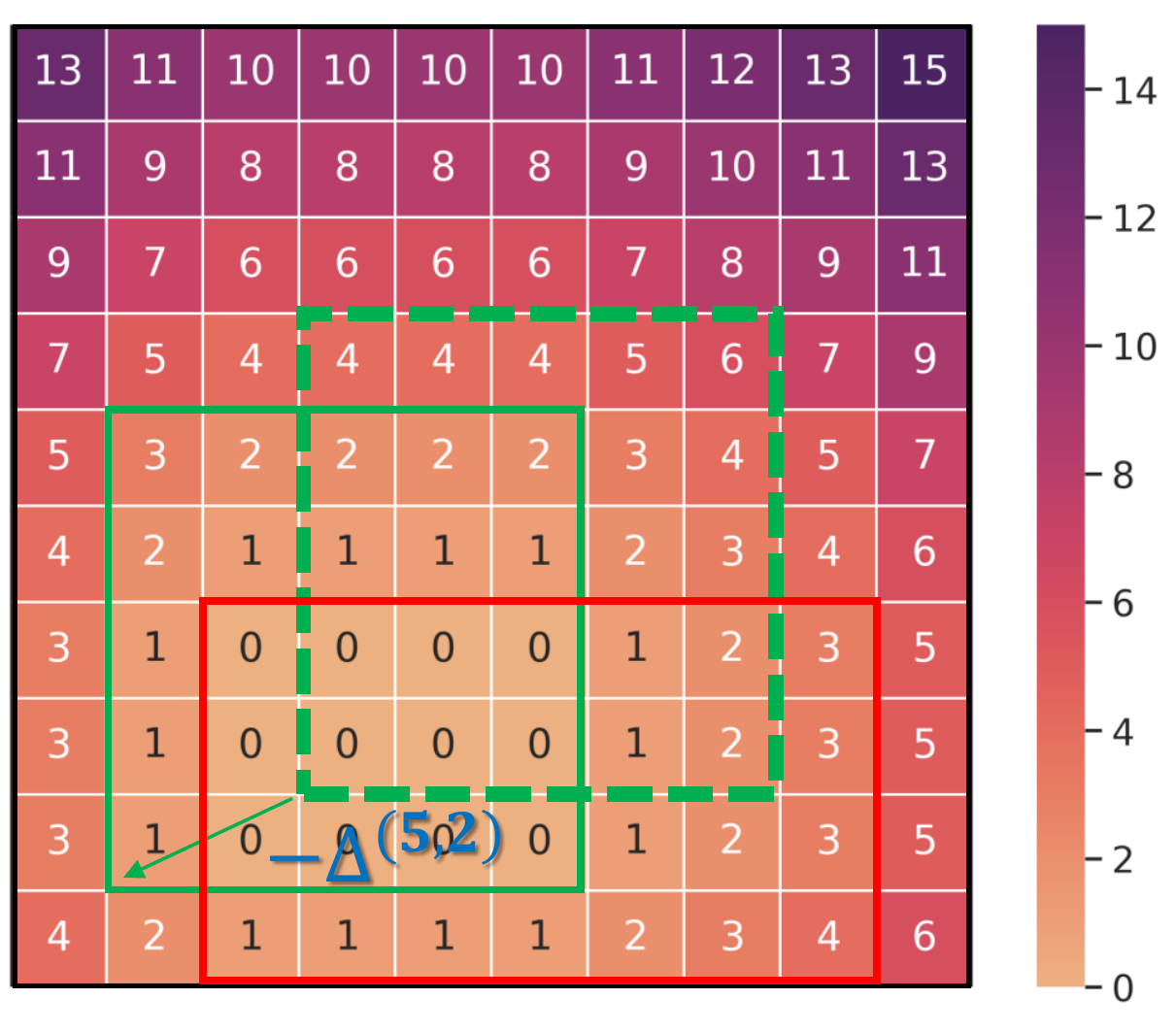}
  \end{center}
  \caption{\small{Wire Mask Example.}}
  \label{wiremask}
\end{wrapfigure}

\textbf{Wire Mask.} 
The wire mask, denoted as $f_w \in [0,1]^{224\times 224}$, is a continuous matrix for representing how HPWL increases if we place a module $M^t$ in a specific position. 
Fig.\ref{wiremask} shows a sample of wire mask, where each value means the increase of HPWL. 
The wire mask aims at finding the best position with the minimum increase of the wirelength.
Intuitively, we can calculate the HPWL  at each canvas position, leading to a complexity of $\mathcal{O}(N^2P)$, where $P$ is the total number of pins.
However, a fast algorithm can be designed by considering the relationships between the pin offset, the net bounding box, and the linear property of the HPWL metric.
For example, Fig.\ref{posmask} illustrates that the next module $M^5$ has two pins, $P^{(5,1)}$ and $P^{(5,2)}$, belonging to ``Net 1'' and ``Net 2'' respectively (Fig.\ref{placementgraph}(e)).
Fig.\ref{wiremask} illustrates the increase of wirelength when placing $M^5$ at each canvas location.
For instance, if $M^5$ is at the bottom-left corner, its Manhattan distance to the two net bounding boxes (in red and green) is $2+2=4$.
To calculate the Manhattan distance more accurately, we move the net bounding box compared to the pin location. For example, since $P^{(5,2)}$ is located at $(2,1)$\footnote{We index the bottom-left corner as the origin $(0,0)$ in a two-dimensional coordinate.}, we move the bounding box of Net 2 (in green) in the direction $-\Delta^{(5,2)}=(-2,-1)$ to encode the information of pin offset.
The time complexity can be reduced to $\mathcal{O}(NP)$. The algorithm can be found in Appendix \ref{algorithm}.

%better positions to place.
% In this algorithm, we need to maintain net bounding boxes during the placement process. 
%an example of wire mask if the next module is $M^5$ (the same as Fig.\ref{posmask}) and the existing placed modules are shown in Fig.\ref{placementgraph}(e).

%, and pin offsets are $(0,0)$ and $(2,1)$.
%We move the net bounding boxes in the opposite directions of the pin offsets $-(\Delta_x, \Delta_y)$.
% according to the definition of HPWL.
%So the Net 2 bounding box moves in the direction $(-2,-1)$ as Fig. \ref{wiremask}.
%Then, we generate the extended Manhattan distance matrix for each net and sum them up to get the wirelength mask.
%Each value in this distance matrix means the HPWL increase if we place the module $M^5$ in this position. 
% Then, we sum up all distance matrices
%The time complexity can be reduced to $O(NP)$ in one episode when row/column assignment costs constant time. 

\textbf{View Mask.} The view mask, denoted as $f_v \in \{0, 1\}^{224\times 224}$, is a global observation of the current chip layout, where the value ``$1$'' means a module has occupied this grid cell. Different from DeepPR \cite{cheng2021joint} that assumed all modules have unit size, we consider real sizes of modules. For instance, if a module has size $w \times h$, it covers $\lceil w  N / W \rceil \times \lceil  h  N / H \rceil$ cells in the canvas, where $W$ and $H$ represent the canvas size and $\lceil\cdot\rceil$ denotes the ceiling function. 
%We assign the value $0.5$ to the boundary cells to avoid boundary disappearance.

%We follow the actor-critic diagram \cite{konda1999actor} and PPO2 reinforcement learning framework \cite{schulman2017proximal}, which trains a policy $\pi_{\theta}(a_t|s_t)$ and a value function $V_{\theta}(s_t)$ with parameters $\theta$.
%We regard the view of placed t-1 modules on the chip as the observation $o_t$.
%Although pin-related features of placed modules are not easy to observe, we should also consider them in the state $s_t$.
%We also take the features of module $M^t$ and $M^{t+1}$ to consider the interaction of placement between the following modules.
%So, the state $s_t$ can be recognized as features of $M^{1:t-1}, M^t, M^{t+1}$. Detailed features in status are listed in Tab. \ref{state}.
% We regard the basic information of module $M^t$ and $M^{t+1}$ as the observation $o_t$ to consider the interaction of placement between the following modules.
% We consider previous actions and observation in the state $s_t=(o_t, a_{1:t-1})$.
%Action $a_t$ is the position to place. We divide the chip into $N^2$ grids, so it has $N^2$ possibilities. We define the position of the left-bottom grid of the module as the action if it occupies more than one grid.

%The architecture of our model as Fig. \ref{architecture} contains three main parts: pixel mask generation, policy network, and value network.

\textbf{Learning Algorithm.}
We train different blocks in Fig.\ref{architecture} as a whole using reinforcement learning. The detailed network architectures are provided in Appendix \ref{detailarch}.
Firstly, we apply the above masks to represent the entire circuits and feed them to downstream networks. 
Secondly, a global feature encoder embeds the view mask of current placement and the wire masks of the following two steps into an embedding vector. Then we combine it with the positional embedding of the $t$-th circuit module in the value network to generate a scalar to evaluate the current state by fully-connected layers. 
Thirdly, a global mask decoder recovers a feature map of size $N^2$, which is fused with different position masks and wire masks in the policy network using $1\times1$ convolutions to avoid the local signal diffusion. The policy network predicts a probability action matrix of size $N\times N$, indicating where to put the next module.
Before sampling actions, we can remove unfeasible actions using the position mask.
Finally, the congestion satisfaction block applies the congestion threshold on the probability matrix to select a final action.

% Secondly, a global feature encoder embeds the view mask of current placement and the wire mask of the next two steps into an embedding vector, which is then combined with the positional embedding of the $t$-th circuit module in the value network to generate a scalar to evaluate the current state by fully-connected layers. 

%In detail, it converts the information of next two modules $M^t, M^{t+1}$ and placed modules $M^{1:t-1}$ (height $M_h$, width $M_w$, pin offset $\Delta$, pin to net connection $P_n$ , and placed position (for placed module only) $M_x, M_y$) to 2 position masks, 2 wirelength masks and 1 view mask.
%  for taking actions.
%The policy network includes , global mask decoder, local mask fusion, merge, and congestion satisfaction modules. The global mask encoder embeds the global features from wirelength masks and the view mask, and then the decoder recovers the same size ($N^2$) matrix. The local mask fusion module accepts position masks and wirelength masks as input and merges information between channels by 1x1 conv to avoid the diffusion of the position signal. Then, global and local features are fused in the merge module to get a probability matrix ($N^2$) for taking actions.Before sampling actions, we remove unfeasible actions by the position mask.The congestion satisfaction module considers the congestion threshold and the probability matrix to give a final placement action in the testing.The value network combines the information of the place index and global embedding to generate a scalar value to evaluate the current state by fully connected layers.

\textbf{Reinforcement Learning.} We borrow the representative actor-critic diagram \cite{konda1999actor} and PPO2 framework \cite{schulman2017proximal} to train the policy $\pi_{\theta}(a_t|s_t)$, where the state representation $s_t$  is listed in Table \ref{state} in Appendix. The action $a_t$ is the canvas position (cell) to place the circuit module. Specifically, we treat  the chip canvas as a grid and divide it into $N\times N$ cells, leading to $N^2$ possible actions. 
The objective function of the policy network can be formulated as
\begin{equation}
L_{\mathrm{policy}}(\theta) = \hat{\mathbb{E}}\Big[\min \big(r_t(\theta)\hat{A}_t,~ \rm{clip}(r_t(\theta), 1-\epsilon, 1+\epsilon)\hat{A}_t\big) \Big],
\end{equation}
where the ratio $r_t(\theta) = \frac{\pi_\theta(a_t|s_t)}{\pi_{\theta_{\mathrm{old}}}(a_t|s_t)}$ and $\hat{A}_t = G_t - \hat{V}_t$ denotes the advantage function. We employ $G_t=\sum_{k=0}^{V-t-1}\gamma^kr_{t+k+1}$ that is the cumulative discounted reward and $\hat{V}_t$ is the estimated value produced by the value network.
We update the the value network by optimizing its objective, 
$L_{\mathrm{value}}(\phi) = \hat{\mathbb{E}}\big[(G_t - \hat{V}_t)^2\big]$.
%We define the position of the left-bottom grid of the module as the action if it occupies more than one grid.
% difference between discounted reward sum and estimated value.
%
% We fix the parameter of global mask encoder when updating value network
% to avoid the effect on policy network.
% We regard the basic information of module $M^t$ and $M^{t+1}$ as the observation $o_t$ to consider the interaction of placement between the following modules.
% We consider previous actions and observation in the state $s_t=(o_t, a_{1:t-1})$.

\textbf{Reward $r_t$.}
We treat HPWL as the reward because wirelength is the main optimization target in different performance metrics. This is different from prior arts \cite{mirhoseini2021graph, cheng2021joint} that weighted combines HPWL and congestion as the reward, which introduces the weighting coefficient as an extra hyper-parameter to tune.
Specifically, we achieve a dense reward by defining a partial HPWL, which only computes the currently placed pins. 
For example, the partial HPWL for $t$ modules can be defined as $\mathrm{HPWL}_t$. 
In other words, we compute $\mathrm{HPWL}_t$ after taking action $a_t$. 
The reward for the step $t$ is $r_t=\mathrm{HPWL}_{t-1}-\mathrm{HPWL}_{t}$, \ie the opposite number of the increase of HPWL. 
Furthermore, instead of computing HPWL at each step, we can maintain the ranges of all net bounding boxes in one episode and update the changes of their sizes with a cost of $\mathcal{O}(P)$, where $P$ is the number of pins. 
Thus we can generate dense rewards while maintaining efficiency.

\textbf{Training and Testing.} Before training, we follow previous work \cite{mirhoseini2021graph} to sort the circuit modules according to the number of nets, areas, and the number of connected modules that have been placed to determine the place order. In training, we update the policy and value networks at each epoch by ignoring the congestion satisfaction block. When updating the value network, we stop the gradient back-propagated in the global mask encoder to avoid influence on the policy network. The detailed training setup is provided in Appendix \ref{detailtrain}.

In the testing stage, for each step $t$, we obtain a probability matrix from the policy network and sample one place action $a_t$.
Then, the congestion satisfaction block will check whether the congestion exceeds a threshold $C_{\mathrm{th}}$ after applying this action. 
If it exceeds, we uniformly sample a few actions, look up the corresponding values from the wire mask $f_w^t$, and estimate the congestion before taking these actions. 
We choose the action with the minimal value in $f_w^t$ and the congestion lower than $C_{\mathrm{th}}$.
If all actions cannot satisfy congestion $C_{\mathrm{th}}$, we select the action with the minimal congestion and move to the next step.
Detailed congestion satisfaction algorithm can be seen in Appendix \ref{algorithm}.

\section{Experiments}
\label{Evaluation}

We extensively evaluate MaskPlace and compare it with several recent advanced placement methods, including NTUPlace3 \cite{chen2008ntuplace3}, RePlAce \cite{cheng2018replace}, DREAMPlace \cite{lin2020dreamplace}, Graph Placement \cite{mirhoseini2021graph}, and DeepPR \cite{cheng2021joint}, where NTUPlace3, RePlAce and DREAMPlace are optimization-based methods, whilst Graph Placement and DeepPR are learning-based approaches. All of them are evaluated on different public circuit benchmarks. All previous works are evaluated by following their experimental settings.

%\subsection{Benchmark}
\textbf{Benchmark.} 
We evaluate MaskPlace in 24 circuit benchmarks 
selected from public datasets  including the widely-used ISPD2005 \cite{nam2005ispd2005}, IBM benchmark suite \cite{adya2009iccad}, and Ariane RISC-V CPU design \cite{zaruba2019cost}. 
The number of evaluated benchmarks is three times more than previous work  \cite{lin2020dreamplace, cheng2021joint, mirhoseini2021graph}.
The statistics of benchmarks are given in Table \ref{benchmark} in Appendix \ref{morebench}, where the largest circuit contains 1,293 macros, 22,802 pins, and more than a million standard cells, leading to a vast state space as aforementioned.
%  (\pluo{can put table in appendix if space is not enough. Add all circuits in ibm benchmark to the statistics table. Modify the description in the paper accordingly})
%These datasets  clustered the standard cells as soft macros.  Detailed statistics are in Tab. \ref{benchmark}. Hard macros in the table are part of macros which are placed by RL in graph placement \cite{mirhoseini2021graph} and other soft macros are placed by traditional placement method. 
%The number of nets, pins and area util are macros. Ports can be seen as fixed and no-size modules.
%\subsection{Experiment Results}
%(such as ``adaptec1'', ``adaptec2'', ``adaptec3'', ``adaptec4'', ``bigblue1'', ``bigblue3'', and ``ariane'') 

\textbf{Main Results.} 
Table \ref{HPWLmacro} compares the HPWL results between all the above methods to place all macros. %
To enable a fair comparison, we evaluate all approaches\footnote{The random seed does not apply in a classic method NTUPlace3.} by using five random seeds and report the means and variances. 
Since the original DeepPR method did not capture macro size (thus does not avoid overlap between adjacent macros because all macros have unit size), we extend DeepPR to model macro size to reduce the overlapping ratio. We name it ``DeepPR-no-overlap''.
Similar to prior works, we use the minimum spanning tree algorithm \cite{cormen2022introduction} to estimate routing wirelength \cite{liao2020attention}. From Table \ref{HPWLmacro}, we can see that MaskPlace achieves the lowest wirelength in six out of seven benchmarks (except ``adaptec4'' where it still outperforms all learning-centric methods). We also see that the conventional optimization-based approaches may fail when 
the circuit benchmark has high chip area usage, such as ``bigblue3 '' and ``ariane''.
Also, MaskPlace gets the lowest wirelength compared with Graph Placement and simulated annealing \cite{kirkpatrick1983optimization} in the IBM benchmark, which is shown in Appendix \ref{suppexp}.
This project website\footnote{\href{https://laiyao1.github.io/maskplace/}{laiyao1.github.io/maskplace}} visualizes and compares different placements.

\begin{table}[!ht]
	\caption{\small{Comparisons of HPWL ($\times 10^5$). HPWL is the smaller the better. We see that MaskPlace outperforms other methods by large margins in six out of seven benchmarks. The traditional optimization such as NTUPlace3 and DREAMPlace may fail in a few benchmarks such as ``ariane''.}}
	\label{HPWLmacro}
    \centering
    % \resizebox{\textwidth}{13mm}{
    \begin{threeparttable}
    \begin{adjustbox}{max width=0.98\textwidth}
    \begin{tabular}{cccccccc}
    
    \toprule
        Method & adaptec1 & adaptec2 & adaptec3 & adaptec4 & bigblue1 & bigblue3 & ariane \\ \midrule
        Random & 61.00±3.85 & 483.12±13.65 & 576.25±16.03 & 600.07±14.17 & 36.67±3.18 & 918.05±43.49 & 52.20±0.90 \\ 
        NTUPlace3 \cite{chen2008ntuplace3}& 26.62 & 321.17 & 328.44 & 462.93 & 22.85 & 455.53 & LG fail\\
        RePlAce \cite{cheng2018replace} & 16.19±2.10 & 153.26±29.01 & 111.21±11.69 & 37.64±1.05 & 2.45±0.06 & 119.84±34.43 & 
        LG fail\\
        DREAMPlace \cite{lin2020dreamplace}& 15.81±1.64 & 140.79±26.73\tnote{1} & 121.94±25.05 & \textbf{37.41±0.87} & 2.44±0.06 &107.19±29.91\tnote{2} & LG fail  \\
        Graph Placement \cite{mirhoseini2021graph}& 30.10±2.98 & 351.71±38.20 & 358.18±13.95 & 151.42±9.72 & 10.58±1.29 & 357.48±47.83 & 16.89±0.60 \\ 
        DeepPR \cite{cheng2021joint} & 19.91±2.13 & 203.51±6.27 & 347.16±4.32 & 311.86±56.74 & 23.33±3.65 & 430.48±12.18 & 52.20±0.89\\
        DeepPR-no-overlap\ \cite{cheng2021joint} & 47.39±4.02 & 425.86±19.59 & 545.40±16.40 & 525.51±10.85 & 26.29±1.48 & 815.10±40.36 & 62.82±0.82 \\ 
        %MaskPlace-Greedy & 8.33±0.79 & 95.57±9.41 & 92.65±6.72 & 91.89±3.93 & 4.22±0.73 & 172.91±10.41 & 18.47±2.02  \\ 
        MaskPlace & \textbf{6.38±0.35} & \textbf{73.75±6.35} & \textbf{84.44±3.60} & 79.21±0.65 & \textbf{2.39±0.05} & \textbf{91.11±7.83} & \textbf{14.63±0.20}  \\ 
        % \midrule
        % DeepPR  (Routing) & ~ & ~ & ~ & ~ & ~ &   \\ 
        % MaskPlace (Routing) & ~ & ~ & ~ & ~ & ~ &   \\ 
        \bottomrule
    \end{tabular}
    \end{adjustbox}
    \begin{tablenotes}
    % \item[*] We modify the raw DeepPR method to take real size of macros into consideration and avoid overlaps. 
    \scriptsize
    \item[1] 2 (of 5) seeds fail in legalization (LG).
    \item[2] 1 (of 5) seed fails in legalization (LG).
    \end{tablenotes}
    \end{threeparttable}
    % \caption{HPWL results of methods}
    % }
\end{table}

\textbf{Compare to Graph Representation.} Since Graph Placement \cite{mirhoseini2021graph} is the recent advanced learning-based approach that employs hypergraph for placement, we compare MaskPlace with it in all four performance metrics, including HPWL, congestion, density, and overlap area ratio. Table \ref{fullmetricsariane} and \ref{fullmetricsispd} report the results. The overlap area ratio describes the ratio of the overlapping area between macros divided by the chip area. 
In Table \ref{fullmetricsariane}, MaskPlace (soft constraint) means that the round function rather than the ceiling function is used to calculate the number of grid cells occupied by the placed macros. MaskPlace with soft constraints may produce better HPWL and congestion than its counterpart with hard constraints, but the overlap area ratio could not be zero because the constraints have been relaxed.  
From Table \ref{fullmetricsariane} and \ref{fullmetricsispd}, we see that MaskPlace outperforms graph placement by large margins, especially in the ISPD benchmark, where the former reduces HPWL compared to the latter one by up to 80\% while ensuring zero overlaps in all benchmarks.
More results in the IBM benchmark can be found in Appendix Table \ref{ibm}. 

\begin{table}[!ht]
     \scriptsize
	\caption{\small{Comparisons between GraphPlace \cite{mirhoseini2021graph} and the proposed MaskPlace using different performance metrics (normalized to $[0,1]$) in the ``ariane'' benchmark, including HPWL, congestion, density, and overlap area ratio. All values are smaller the better.  We see that MaskPlace surpasses other methods significantly while ensuring zero overlaps, which is essential for a valid and manufacturable chip layout.}}
	\label{fullmetricsariane}
    \centering
    \begin{tabular}{cccccc}
    \toprule
        Method & HPWL & Congestion & Density & Overlap (\%) \\ \midrule
        Graph Placement (journal) \cite{mirhoseini2021graph} & 0.1198±0.0019 & 0.9718±0.0346 & 0.5729±0.0086 & 5.13±0.11  &   \\ 
        Graph Placement (github) \cite{mirhoseini2021graph} & 0.1013±0.0036 & 0.9174±0.0647 & 0.5502±0.0568 & 4.29±1.25 & \\ 
        MaskPlace (hard constraint) & 0.1025±0.0015 & 1.0137±0.0451 & \textbf{0.5000±0.0000} & \textbf{0.00±0.00} &  \\ 
        MaskPlace (soft constraint) & \textbf{0.0879±0.0012} & \textbf{0.9049±0.0115} & 0.5262±0.0015 & 3.33±0.79 &   \\ 
        \bottomrule
    \end{tabular}
\end{table}

\begin{table}[!ht]
     \scriptsize
	\caption{\small{Comparisons between GraphPlace \cite{mirhoseini2021graph} and MaskPlace in four performance  metrics (normalized to $[0,1]$) in the ISPD benchmark. All
values are smaller the better. We see that MaskPlace can reduce the HPWL up to 80\% compared to its counterpart while ensuring the modules' overlaps are zeros in all benchmarks.}}
	\label{fullmetricsispd}
    \centering
    \begin{tabular}{ccccccccc}
    \toprule
    \multirow{2}{*}{benchmark} &\multicolumn{4}{c}{Graph Placement \cite{mirhoseini2021graph}} &  \multicolumn{4}{c}{MaskPlace} \\
    \cmidrule(r){2-5} \cmidrule(r){6-9}
        ~ & HPWL & Congestion & Density & Overlap(\%) & HPWL & Congestion & Density & Overlap (\%)  \\ \midrule
        adaptec1 & 0.1810 & 0.7370 & 0.5340 & 1.89 & \textbf{0.0384} & \textbf{0.6961} & \textbf{0.5000} & \textbf{0.00}   \\ 
        adaptec2 & 0.2814 & 0.7387 & 0.5147 & 1.54 & \textbf{0.0549} & \textbf{0.6990} & \textbf{0.5000} & \textbf{0.00}   \\ 
        adaptec3 & 0.2248 & 0.7431 & 0.5226 & 1.24 & \textbf{0.0540} & \textbf{0.7130} & \textbf{0.5000} & \textbf{0.00}   \\ 
        adaptec4 & 0.1107 & 0.7369 & 0.7472 & 7.59 & \textbf{0.0560} & \textbf{0.7078} & \textbf{0.5000} & \textbf{0.00}   \\ 
        bigblue1 & 0.0958 & 0.7346 & 0.5181 & 1.98 & \textbf{0.0255} & \textbf{0.6953} & \textbf{0.4876} & \textbf{0.00}   \\ 
        bigblue3 & 0.1565 & 0.7499 & 0.5174 & 0.96 & \textbf{0.0430} & \textbf{0.7350} & \textbf{0.5000} & \textbf{0.00}   \\ 
        
        \bottomrule
    \end{tabular}
\end{table}

% \pluo{add a table of statistics for the IBM benchmark in appendix.}.

% $\lceil width \times N / max_{width} -1/2\rceil \times \lceil height \times N / max_{height} -1/2\rceil$

%Because only graph placement \cite{mirhoseini2021graph} takes all three metrics, we compare with it in them. The results are in Tab. \ref{fullmetricsariane} and Tab. \ref{fullmetricsispd}. Also, we introduce another metric overlap area to describe the ratio of the overlapping area to the chip area. 

\textbf{Routing Wirelength.} Table \ref{routingtab} compares the routing wirelength between MaskPlace and DeepPR \cite{cheng2021joint}. We show that using the true wirelength as the reward would lower the efficiency and produce a sparse reward. We see that MaskPlace, which employs HPWL as the reward, can achieve 60\% to 90\% shorter routing wirelength than DeepPR, which directly used the true wirelength as the reward.

\begin{table}[!ht]
     \scriptsize
	\caption{\small{Compare routing wirelength  ($\times 10^5$)  between DeepPR \cite{cheng2021joint} and MaskPlace.}}
	\label{routingtab}
    \centering
    \begin{adjustbox}{max width=0.98\textwidth}
    \begin{tabular}{cccccccc}
    \toprule
        method & adaptec1 & adaptec2 & adaptec3 & adaptec4 & bigblue1 & bigblue3 & ariane \\ \midrule
        DeepPR \cite{cheng2021joint} & 23.25±3.03 & 212.97±5.84 & 377.80±5.49 & 367.57±64.44 & 28.51±3.90 & 507.39±14.82  & 56.77±0.87\\ 
        DeepPR-no-overlap \cite{cheng2021joint} & 52.46±3.97 & 451.22±19.00 & 583.32±15.92 & 628.22±10.02 & 31.02±1.41 & 945.60±43.24  & 68.89±0.81\\ 
        MaskPlace & \textbf{7.12±0.34} &  \textbf{77.70±6.77} & \textbf{90.40±3.82} & \textbf{92.51±0.38} & \textbf{2.81±0.51} & \textbf{103.24±10.48} & \textbf{15.61±0.19} \\ 
        \bottomrule
    \end{tabular}
    \end{adjustbox}
\end{table}

\textbf{Standard Cells.} Table \ref{standardcell} compares the HPWL of both the macros and the standard cells by using MaskPlace, DeepPR \cite{cheng2021joint}, and DREAMPlace \cite{lin2020dreamplace}, where DREAMPlace is employed to place the standard cells following the experimental setup in \cite{cheng2021joint}. We can see that the proposed method surpasses the other approaches by up to 50\% in the large benchmark ``bigblue3'', which has more than a million standard cells. 
%, showing that MaskPlace can be applied to all module placement. 
% \begin{table}[htbp]
%      \small
% 	\caption{Congestion Search}
%     \centering
%     \begin{tabular}{ccccc}
%     \toprule
%         Pre-defined $C_{th}$ & DREAMPlace & Graph Placement & DeepPR & MaskPlace  \\ \midrule
%         infer time \\
%         infer time (per macro)\\
%     \bottomrule
%     \end{tabular}
% \end{table}

\begin{table}[!ht]
     \scriptsize
	\caption{\small{Comparisons of HPWL ($\times 10^7$) for macro and standard cell placement.}}
	\label{standardcell}
    \centering
    \begin{tabular}{ccccccc}
    \toprule
        Method & adaptec1 & adaptec2 & adaptec3 & adaptec4 & bigblue1 & bigblue3  \\ \midrule
        DREAMPlace \cite{lin2020dreamplace} & 11.01±1.37& 16.19±2.60& 21.54±1.19 &35.47±4.97& 10.28±1.11 &70.02±46.06\\
        DeepPR \cite{cheng2021joint} + DREAMPlace \cite{lin2020dreamplace} & 8.01 & 12.32 & 24.11 & 23.64 & 14.04 & 45.06  \\ 
        MaskPlace + DREAMPlace \cite{lin2020dreamplace} & \textbf{7.93±0.20} & \textbf{9.95±0.29} & \textbf{21.49±0.90} & \textbf{22.97±0.92} & \textbf{9.43±0.13} & \textbf{37.29±0.67}  \\ 
        \bottomrule
    \end{tabular}
\end{table}

\paragraph{Placement w/o real size}
Considering that DeepPR ignored the module size, we implement MaskPlace in the same setting, and the result can be found in Table \ref{norealsize}. The result shows that our method has significant advantages.

\begin{table}[!ht]
 \scriptsize
	\caption{Routing wirelength for macro placement w/o real size}
	\label{norealsize}
    \centering
    \begin{tabular}{ccccccc}
    \toprule
        Method & adaptec1 & adaptec2 & adaptec3 & adaptec4 & bigblue1 & bigblue3  \\ \midrule
        DeepPR \cite{cheng2021joint} & 5298 & 22256 & 32839 & 63560 & 8602 & 94083  \\ 
        MaskPlace & \textbf{2941} & \textbf{20593} & \textbf{16181} & \textbf{18553} & \textbf{2331} & \textbf{27403}  \\ 
        \bottomrule
    \end{tabular}
\end{table}

\paragraph{Transferability}
Test the performance of the model trained on \textit{adaptec1} on other benchmarks as Table \ref{transfer}. The results show that our method also has a good transferability.

\begin{table}[!ht]
 \scriptsize
	\caption{\small{HPWL ($\times 10^5$) results for transferability. HPWL is the smaller the better. The model has been trained on \textit{adaptec1} benchmark and just took the inference in other benchmarks.}}
	\label{transfer}
    \centering
    % \resizebox{\textwidth}{13mm}{
    % \begin{adjustbox}{max width=0.98\textwidth}
    \begin{threeparttable}
    {
    \begin{tabular}{cccccccc}
    
    \toprule
         & adaptec2 & adaptec3 & adaptec4 & bigblue1 & bigblue3 & ariane \\ \midrule
        HPWL & 85.56±9.41 & 89.77±6.72 & 87.32±3.93 & 2.87±0.31 & 160.63±10.41 & 19.32±2.02 \\ 
       ratio\tnote{*} & 1.16 & 1.06 & 1.11 & 1.20 & 1.76 & 1.32\\ 
        % \midrule
        % DeepPR  (Routing) & ~ & ~ & ~ & ~ & ~ &   \\ 
        % MaskPlace (Routing) & ~ & ~ & ~ & ~ & ~ &   \\ 
        \bottomrule
    \end{tabular}
    }
    \begin{tablenotes}
    % \item[*] We modify the raw DeepPR method to take real size of macros into consideration and avoid overlaps. 
    \small
    \item[*] Compared with the result from the model trained on the corresponding benchmark.
    \end{tablenotes}
    \end{threeparttable}
    % \end{adjustbox}
    % \caption{HPWL results of methods}
    % }
\end{table}

\textbf{Efficiency.} Table \ref{efficiency} compares the inference efficiency of different approaches. All of them are evaluated on one GeForce RTX 3090 GPU, and the CPU version of DREAMPlace is allocated with 16 threads in a 16 CPU cores environment. We see that the careful design of MaskPlace makes it faster than two other learning-based approaches.

\begin{table}[!ht]
	\caption{\small{Comparisons of wall-clock runtime (second) of different placement methods in inference.}}
	\label{efficiency}
	\scriptsize
    \centering
    \begin{tabular}{ccccccc}
    \toprule
        Method & adaptec1 & adaptec2 & adaptec3 & adaptec4 & bigblue1 & bigblue3 \\ \midrule
        DREAMPlace (CPU) \cite{lin2020dreamplace} & 4.47  & 11.50  & 11.52  & 15.55  & 9.32  & 27.36 \\ 
        DREAMPlace (GPU) \cite{lin2020dreamplace} & 4.51  & 7.57  & 7.70  & \textbf{7.39}  & 5.57  & \textbf{12.25} \\ 
        Graph Placement \cite{mirhoseini2021graph} & 6.32  & 16.97  & 20.05  & 13.40  & 4.54  & 15.65 \\ 
        DeepPR \cite{cheng2021joint} & 10.25  & 10.46  & 22.82  & 42.24  & 9.86  & 32.53 \\ 
  MaskPlace & \textbf{4.26}  & \textbf{6.98 } & \textbf{7.63}  & 13.36  & \textbf{4.32}  & 13.87 \\ 
    \bottomrule
    \end{tabular}
\vspace{-8pt}
\end{table}

\begin{figure}[!ht]
\centering
\begin{minipage}{.49\textwidth}
  \centering
  \includegraphics[width=.97\linewidth]{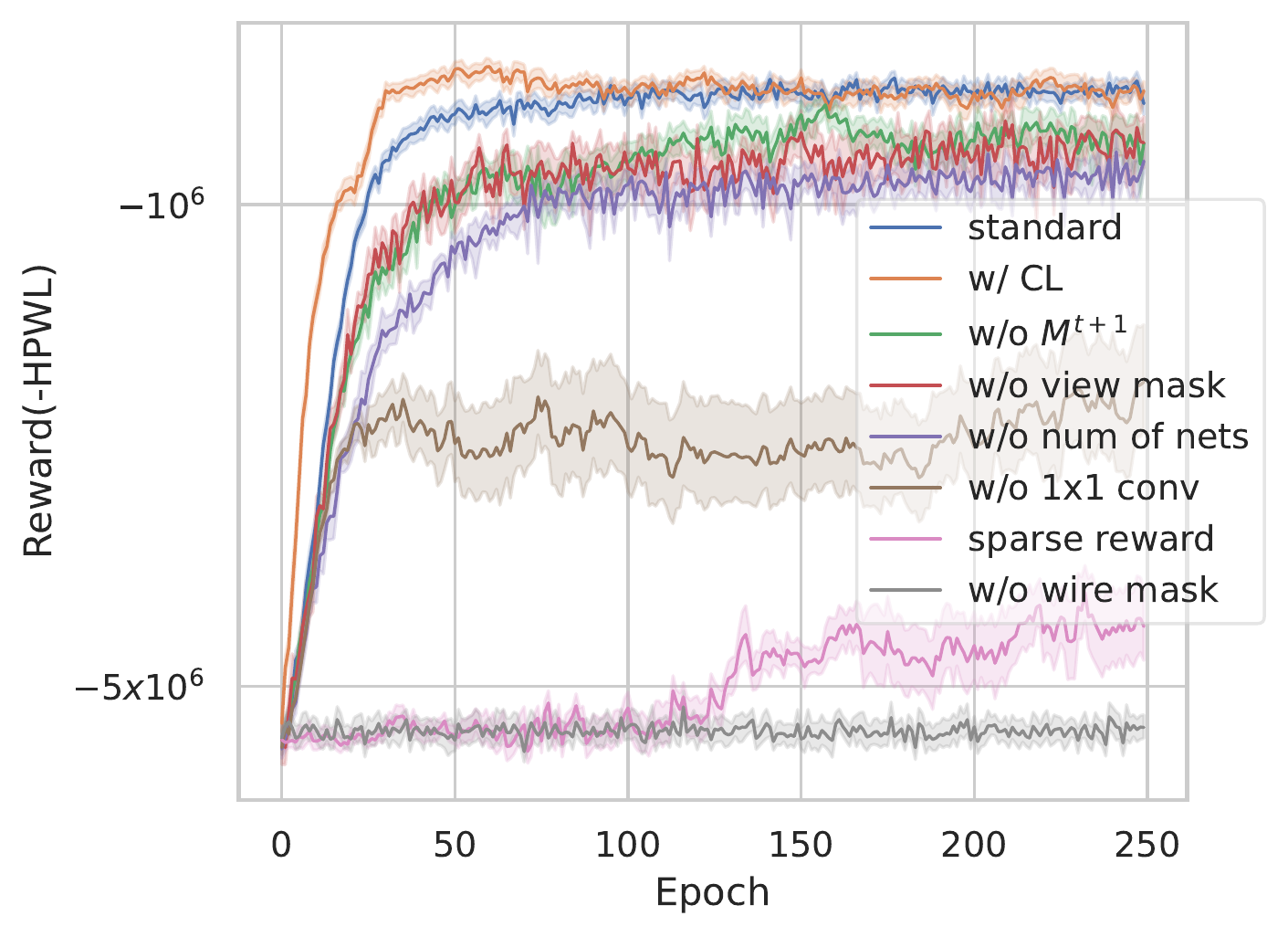}
  \captionof{figure}{\small{Compare reward curves of different components in MaskPlace.}}
  \label{ablation}
\end{minipage}%
\hspace{5mm}
\begin{minipage}{.45\textwidth}
  \centering
  \includegraphics[width=.97\linewidth]{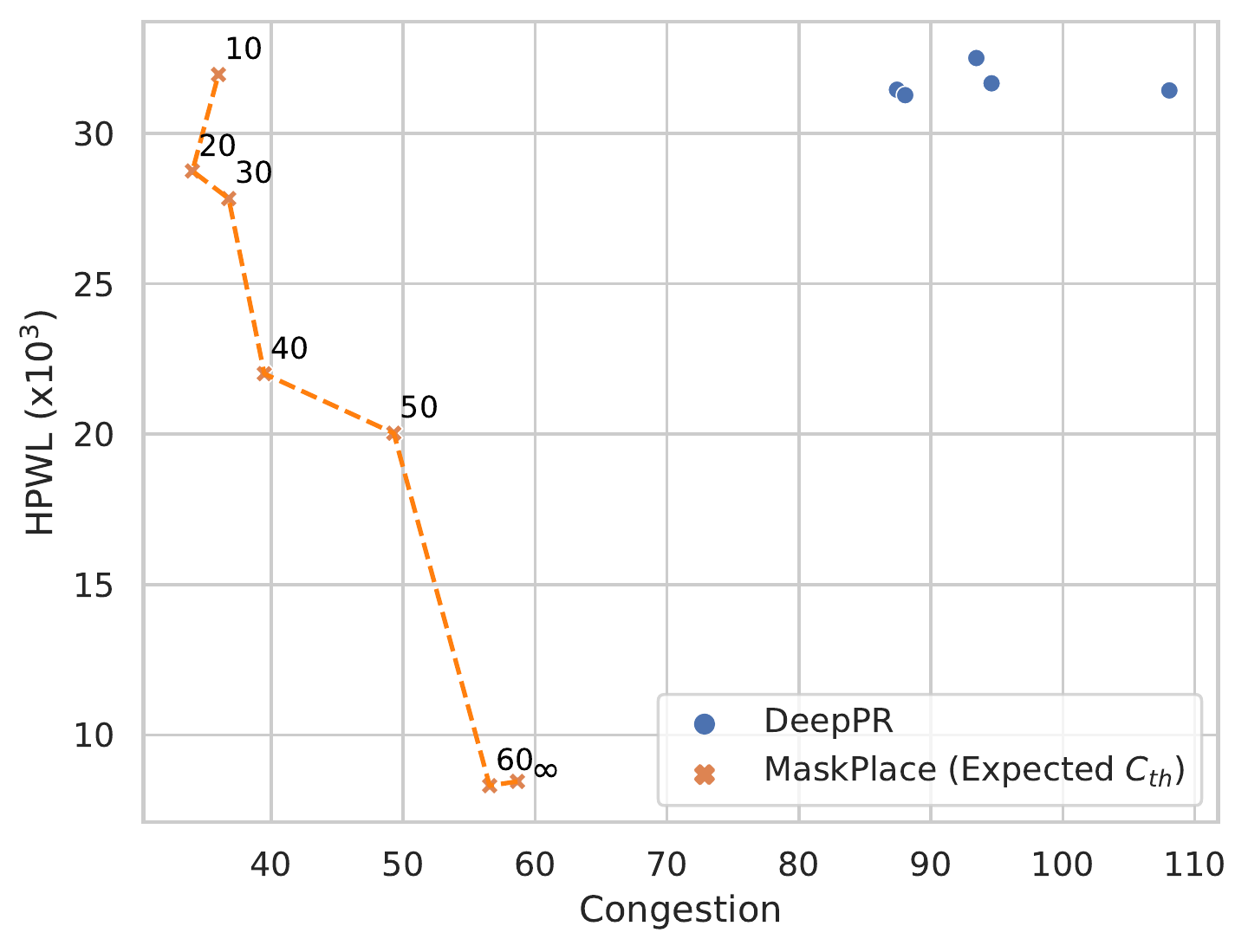}
  \captionof{figure}{Study of congestion satisfaction. }
  \label{congestion}
\end{minipage}
\vspace{-10pt}
\end{figure}

\textbf{Ablation Study.}
We compare different components in MaskPlace as shown in Fig.\ref{ablation}. Each curve is produced by five seeds using the benchmark ``adaptec1''. For example, MaskPlace \textit{w/ CL} means using 1/3 of the circuit macros to pretrain the model for 30 epochs like curriculum learning.
MaskPlace \textit{w/o} $M^{t+1}$ means only considering $M^{t}$ as input without looking one step forward.
Moreover, MaskPlace \textit{w/o number of nets} means this feature is not considered when determining the placement order. 
MaskPlace \textit{w/o 1x1 conv} means that 7x7 kernels are used to replace the 1x1 kernels in the local feature fusion block.
Also, MaskPlace with \textit{sparse reward} means compute HPWL reward only when all macros have been placed, and the rewards for other steps are set to zeros.
MaskPlace \textit{w/o view mask} and \textit{w/o wire mask} means that the corresponding mask is not inputted into the model.
We can see that MaskPlace (standard) with curriculum learning has the best performance.

%The result of ablation study is in Fig. \ref{ablation} with 5 seeds in benchmark \textit{adaptec1}.
%\textit{w/ curriculum learning} means using 1/3 modules to pretrain model for 30 epochs.
%\textit{w/o module} $M^{t+1}$ means only considering the module $M^{t}$ for actions.
%\textit{w/o number of nets} means this feature is not considered in the placement order.   

% \begin{figure}
% 	\centering
% 	\includegraphics[width=0.5\columnwidth]{test.png}
% % 	\fbox{\rule[-.5cm]{4cm}{4cm} \rule[-.5cm]{4cm}{0cm}}
% 	\caption{Reward curves of ablations}
% 	\label{fig:ablations}
% \end{figure}
% % \paragraph{Transferable experiments}
% % \paragraph{Congestion Search}

% \begin{figure}
% 	\centering
% 	\includegraphics[width=0.5\columnwidth]{scatter.png}
% % 	\fbox{\rule[-.5cm]{4cm}{4cm} \rule[-.5cm]{4cm}{0cm}}
% 	\caption{Congestion Search}
% 	\label{fig:congestion}
% \end{figure}
% \paragraph{Transferable experiments}
% \paragraph{Congestion Search}

\textbf{Congestion Satisfaction.}
To evaluate our congestion satisfaction block, we implement a placement without any congestion threshold (\ie $\infty$) as shown in Fig.\ref{congestion}. We evaluate the ``adaptec3'' benchmark, where MaskPlace outperforms DeepPR. We gradually lower the threshold $C_{th}$ from 60 to 10. We find that lower congestion leads to an increase in the HPWL. 
Our method can always satisfy the congestion constraint in five seeds in a suitable range (above 40 in this benchmark).
If we continue to reduce the congestion threshold after a specific value (say $40$ in Fig.\ref{congestion}), we found that it hardly satisfies the threshold because nets must take up a certain amount of wire resources.
%We also plot the results of 5 seeds by DeepPR as a comparison.

\section{Conclusion}
\label{Conclusion}
This paper proposes MaskPlace, an RL-based placement method based on rich visual representation by learning position, wirelength, and view information. It helps the model take action effectively and efficiently without reducing the search space. We design a direct reward function based on practical scenarios and get satisfactory results on all key metrics. This work can facilitate the placement process and avoid undesired overlaps between modules.
In the future, we will explore the standard cell placement by designing a suitable representation, which is an open problem for RL due to its vast space.

\textbf{Limitation and Potential Negative Societal Impact.} 
Chip design flow contains many stages, and our method shows its potential in a single stage.
Similar to previous RL methods, it also requires an optimization method when placing millions of standard cells because RL's state space is too large.
Our method does not have potential harm to the public society at the moment.

\begin{ack}
We thank Xibo Sun answering questions about EDA. We also thank Runjian Chen for participating in our discussion. Ping Luo is supported by the General Research Fund of HK No.27208720, No.17212120, and No.17200622.
% Use unnumbered first level headings for the acknowledgments. All acknowledgments
% More information about this disclosure can be found at: \url{https://neurips.cc/Conferences/2022/PaperInformation/FundingDisclosure}.

% Do {\bf not} include this section in the anonymized submission, only in the final paper. You can use the \texttt{ack} environment provided in the style file to autmoatically hide this section in the anonymized submission.
\end{ack}

% \section*{References}

% References follow the acknowledgments. Use unnumbered first-level heading for
% the references. Any choice of citation style is acceptable as long as you are
% consistent. It is permissible to reduce the font size to \verb+small+ (9 point)
% when listing the references.
% Note that the Reference section does not count towards the page limit.
% \medskip

{
\small

\bibliography{references}
% \printbibliography

% [1] Alexander, J.A.\ \& Mozer, M.C.\ (1995) Template-based algorithms for
% connectionist rule extraction. In G.\ Tesauro, D.S.\ Touretzky and T.K.\ Leen
% (eds.), {\it Advances in Neural Information Processing Systems 7},
% pp.\ 609--616. Cambridge, MA: MIT Press.

% [2] Bower, J.M.\ \& Beeman, D.\ (1995) {\it The Book of GENESIS: Exploring
%   Realistic Neural Models with the GEneral NEural SImulation System.}  New York:
% TELOS/Springer--Verlag.

% [3] Hasselmo, M.E., Schnell, E.\ \& Barkai, E.\ (1995) Dynamics of learning and
% recall at excitatory recurrent synapses and cholinergic modulation in rat
% hippocampal region CA3. {\it Journal of Neuroscience} {\bf 15}(7):5249-5262.
}

%%%%%%%%%%%%%%%%%%%%%%%%%%%%%%%%%%%%%%%%%%%%%%%%%%%%%%%%%%%%
\section*{Checklist}

%%% BEGIN INSTRUCTIONS %%%
% The checklist follows the references.  Please
% read the checklist guidelines carefully for information on how to answer these
% questions.  For each question, change the default \answerTODO{} to \answerYes{},
% \answerNo{}, or \answerNA{}.  You are strongly encouraged to include a {\bf
% justification to your answer}, either by referencing the appropriate section of
% your paper or providing a brief inline description.  For example:
% \begin{itemize}
%   \item Did you include the license to the code and datasets? \answerYes{See Section~\ref{gen_inst}.}
%   \item Did you include the license to the code and datasets? \answerNo{The code and the data are proprietary.}
%   \item Did you include the license to the code and datasets? \answerNA{}
% \end{itemize}
% Please do not modify the questions and only use the provided macros for your
% answers.  Note that the Checklist section does not count towards the page
% limit.  In your paper, please delete this instructions block and only keep the
% Checklist section heading above along with the questions/answers below.
%%% END INSTRUCTIONS %%%

\begin{enumerate}

\item For all authors...
\begin{enumerate}
  \item Do the main claims made in the abstract and introduction accurately reflect the paper's contributions and scope?
    \answerYes{}
  \item Did you describe the limitations of your work?
    \answerYes{}
  \item Did you discuss any potential negative societal impacts of your work?
    \answerYes{}
  \item Have you read the ethics review guidelines and ensured that your paper conforms to them?
    \answerYes{}
\end{enumerate}

\item If you are including theoretical results...
\begin{enumerate}
  \item Did you state the full set of assumptions of all theoretical results?
    \answerNA{}
        \item Did you include complete proofs of all theoretical results?
    \answerNA{}
\end{enumerate}

\item If you ran experiments...
\begin{enumerate}
  \item Did you include the code, data, and instructions needed to reproduce the main experimental results (either in the supplemental material or as a URL)?
     \answerYes{}
  \item Did you specify all the training details (e.g., data splits, hyperparameters, how they were chosen)?
     \answerYes{}
        \item Did you report error bars (e.g., with respect to the random seed after running experiments multiple times)?
     \answerYes{}
        \item Did you include the total amount of compute and the type of resources used (e.g., type of GPUs, internal cluster, or cloud provider)?
     \answerYes{}
\end{enumerate}

\item If you are using existing assets (e.g., code, data, models) or curating/releasing new assets...
\begin{enumerate}
  \item If your work uses existing assets, did you cite the creators?
    \answerYes{}
  \item Did you mention the license of the assets?
    \answerNA{}
  \item Did you include any new assets either in the supplemental material or as a URL?
    \answerYes{}
  \item Did you discuss whether and how consent was obtained from people whose data you're using/curating?
    \answerNA{}
  \item Did you discuss whether the data you are using/curating contains personally identifiable information or offensive content?
    \answerNA{}
\end{enumerate}

\item If you used crowdsourcing or conducted research with human subjects...
\begin{enumerate}
  \item Did you include the full text of instructions given to participants and screenshots, if applicable?
    \answerNA{}
  \item Did you describe any potential participant risks, with links to Institutional Review Board (IRB) approvals, if applicable?
    \answerNA{}
  \item Did you include the estimated hourly wage paid to participants and the total amount spent on participant compensation?
    \answerNA{}
\end{enumerate}

\end{enumerate}

%%%%%%%%%%%%%%%%%%%%%%%%%%%%%%%%%%%%%%%%%%%%%%%%%%%%%%%%%%%%
\newpage

\appendix

\section{Appendix}

% Optionally include extra information (complete proofs, additional experiments and plots) in the appendix.
% This section will often be part of the supplemental material.

\subsection{Module, Net and Pin}

\begin{figure}[!ht] 
% \captionsetup[subfigure]{justification=centering}
	\centering  %图片全局居中
% 	\vspace{-0.35cm} %设置与上面正文的距离
% 	\subfigtopskip=2pt %设置子图与上面正文或别的内容的距离
% 	\subfigbottomskip=2pt %设置第二行子图与第一行子图的距离，即下面的头与上面的脚的距离
% 	\subfigcapskip=-5pt %设置子图与子标题之间的距离
    \includegraphics[width=0.99\columnwidth]{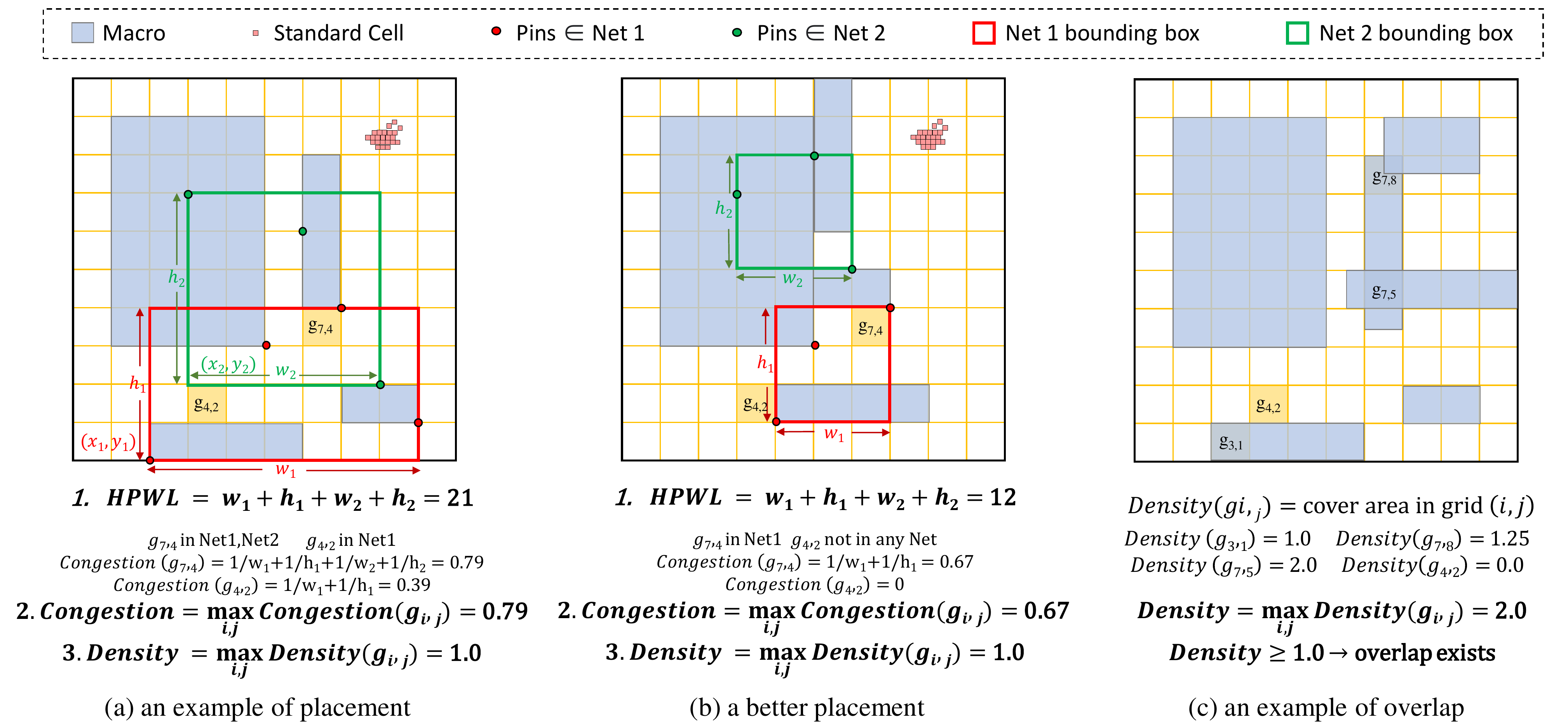}
	
	\caption{\textbf{Metrics for placement.} HPWL is an optimization item, while congestion and density are constraint items in the actual placement scenario. HPWL is smaller the better, while congestion and density need to be less than the given thresholds. Placement (b) is better than (a) because HPWL and congestion of (a) are smaller. Placement (c) is invalid because there are overlaps in cell $g_{7,5}$ and $g_{7,8}$.}
	\label{metrics}
\end{figure}

\paragraph{Module.} A chip is a combination of numerous modules, and there are two types of them: macros and standard cells. Macros are relatively large, including DRAMs, caches, and IO interfaces. Standard cells are mainly logical gates, much smaller than macros, and the size can be ignored. 
As in Fig.\ref{metrics} (a), there are four macros and several standard cells.
Placement methods usually place macros first and then the standard cells to ensure there is enough space for macros \cite{yan2022towards}. 
Due to the considerable number of standard cells, we currently use our MaskPlace method on macro placement.

%  and the size (height and width) of them cannot be ignored
% In many methods, standard cells are be seen as points without size.

\paragraph{Pin.} Pins are input/output interfaces for modules and are connected by wires directly, which have fixed relative positions on modules. We define the relative position of the pin $P^{(i,j)}$ from the left-bottom corner of the module it belongs to as $\Delta^{(i,j)}= (\Delta^{(i,j)}_x, \Delta^{(i,j)}_y)$. For example, there are five pins and three macros in Fig.\ref{modulenetpin} (a), and the pin offset information is also shown at the bottom. 
In the placement task, we should not ignore the positions of pins because it determines the wirelength. However, graph neural network-based models \cite{mirhoseini2021graph, cheng2021joint} ignored them when converting circuits into a graph, which may lead to sub-optimal results.
% Each pin has a fixed offset $(\Delta$ to describe its relative position to the lower left corner of the module. 

\paragraph{Net.} 
A net contains a set of pins connected by the same wires. Thus the pins have the same information (0/1 in digital circuits).
For example, four pins belong to Net 1, and the other three pins belong to Net 2 in Fig.\ref{metrics} (a). Usually, one pin belongs to only one net, and one net has more than two pins (one input and several outputs). Pins from the same net can form a net bounding box as Fig.\ref{metrics} (a)(b).

% and have the same information (0/1 in digital circuits).
% Nets are used to describe the connections between modules. Pins belonging to the same net should be connected with each other.

% We explain why pin offset information is necessary. Suppose we want to minimize wirelength. In the graph based method, the input features for module include size $(w, h)$, place position $(x, y)$ and module type like Fig.\ref{graph}. So, the network can hardly infer the real position of pins and tend to use the center positions of modules to predict the positions of pins. In this way the agent will align the centers of the two modules horizontally and the result of placement will look like Fig.\ref{modulenetpin} (b) to get the wirelength 6. However, when considering the pins are near the bottom of the modules, it is better to align the bottom of the two modules as Fig.\ref{modulenetpin} (c), and thus wirelength will be reduced to 2.

\subsection{Metric}
\label{detailmetric}
\paragraph{HPWL.} HPWL (Half Perimeter Wire Length) is widely used to estimate wirelength by small computation cost \cite{chen2006high}. It is the sum of half perimeter of net bounding boxes as Fig.\ref{metrics} (a)(b), where the bounding box is the minimal rectangle including all pins belonging to this net. 

\paragraph{Congestion.} The congestion metric is used to avoid routing congestion, resulting in an increase in the actual wirelength because the resources for wires are limited in a real chip. 
There are many ways to estimate congestion, one is to compute a rough routing result \cite{mirhoseini2021graph}, but it is very computationally intensive.
We use RUDY \cite{spindler2007fast} as the estimation of congestion, which is a common way to evaluate.
In RUDY, each grid cell needs to accumulate the inverse of the height and width $(1/h+1/w)$ of all the net bounding boxes covering itself and take out the maximum value (or the average of the first k maximums) of all grid cells as Fig.\ref{metrics} (a)(b).

% , which is same as DeepPR\cite{cheng2021joint} used. 
% Routing is a process to determine the precise wire paths for nets. 
% So, if congestion is too high, wires may not be routed optimally, and wirelength will increase. 

\paragraph{Density.}
Density is a metric to reduce overlaps and avoid time-consuming computation for $O(V^2)$ constraints \cite{wang2009electronic}. 
So, it is an approximate calculation essentially.
It is defined as the maximum stackable coverage area ratio for each grid cell on a chip canvas. 
For example, as Fig.\ref{metrics} (c), the maximum stackable coverage area ratio is $2.0$ in grid cell $g_{7,5}$ because two modules fully occupy it. 
However, density less than a small value is not a sufficient condition for the absence of overlap.
Because our method can ensure no overlaps, we only consider it in evaluation. 
In the practical application scenario of chip design, HPWL is an optimization item. Conversely, congestion and density are constraint items. 
% $Density\le1.0$ is only a necessary condition for the absence of overlap. 
% Congestion constraint is for chip routability, and density constraint is for chip feasibility. We do not need to reduce these last two metrics indefinitely. 
% with the number of modules $V$ 
%  for comparison.
\paragraph{Examples.}

We give a set of placement results to explain the metrics in Fig.\ref{metrics}. We can see that HPWL is the sum of width and height of net bounding boxes. Congestion (RUDY) is the max congestion value of grid cell $g_{i,j}$, and the value in each grid cell is cumulative from the reciprocal of the width and height of the net bounding box containing that grid cell. (a) and (b) are from the same circuit, but (b) is a better placement because (b) has lower HPWL and congestion. Density is the max density value of grid cell $g_{i,j}$, and the value in each grid cell is stackable coverage area ratio of the grid cell. The density of Fig.\ref{metrics} (c) is 2.0 because $g_{7,5}$ completely covered by two modules.

\begin{figure}[!ht]
 
  \begin{center}
    \includegraphics[width=0.98\textwidth]{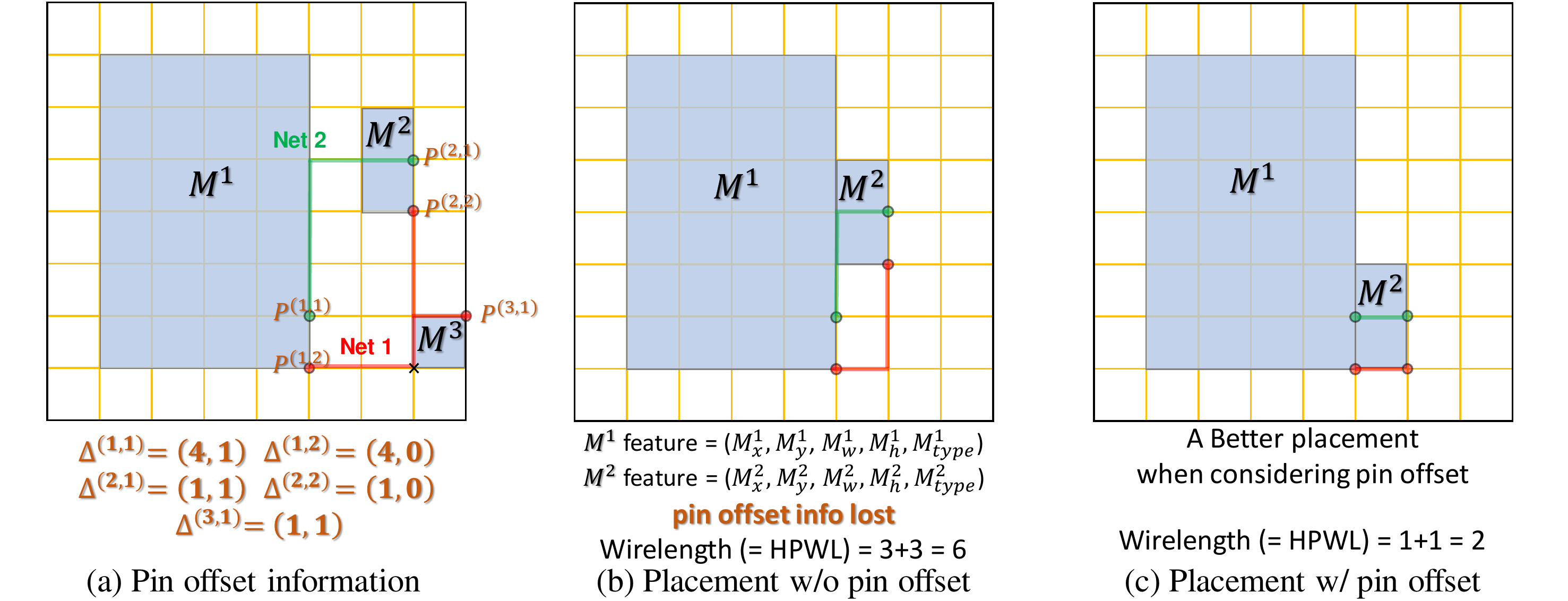}
    
  \end{center}
  \caption{\textbf{Explanation for module, pin and net.} (a) gives an example for pin offset information. When we remove the pin offset information, the model tends to align the centers of the two modules horizontally like (b) because it uses the center position of modules to estimate pin location. However, we have a better design as (c) when considering pins are located on the bottom of the modules.}
  \label{modulenetpin}
\end{figure}

\paragraph{Relationship between pin offset and HPWL.} The pin offset can affect the HPWL. In the graph-based method, the input features for module include size $(M_w, M_h)$, position $(M_x, M_y)$ and type. So, the network can hardly infer the real position of pins and tend to use the center positions of modules to predict the positions of pins. In this way, the agent will align the centers of the two modules horizontally, and the result of placement is like Fig.\ref{modulenetpin} (b) to get the wirelength 6. However, when considering the pins are near the bottom of the modules, it is better to align the bottom of the two modules as Fig.\ref{modulenetpin} (c), and thus wirelength can be reduced to 2 if we consider the pin offset.

\subsection{Algorithms}
\label{algorithm}

\paragraph{Reward Computation.}
The dense reward generation algorithm is shown in Algorithm \ref{algorithm1}. It can generate dense rewards without an efficiency decrease. For simplicity, we omit the calculation of the y dimension, which is the same as the x dimension.

\begin{algorithm}
\caption{Dense HPWL Reward Computation (omit y-dimension)}
\label{algorithm1}
\KwData{Placed position of module $M^t$ $(M_x^t, M_y^t)$, max/min x/y coordinates of nets $MaxMinCoord$,
pin offsets $(\Delta_x^{(t,j)}, \Delta_y^{(t,j)})$, pin to net connection $P_n^{(t,j)}$;}
\KwResult{Incremental HPWL Reward $reward$;}
$reward \leftarrow 0$\;
\ForEach{$\Delta_x^{(t,j)}, P_n^{(t,j)}$ of all pins $P^{(t,j)}$ from $M^{t}$ }{
    $x \leftarrow M_x^t+\Delta_x^{(t,j)}$ \tcp*[l]{\textcolor{blue}{calculate pin coordinates}}\
    \eIf{ $P_n^{(t,j)}$ not in $MaxMinCoord$ }{
        \tcp{\textcolor{blue}{The net for the first time has a definite location of the pin}}
        $MaxMinCoord[P_n^{(t,j)}].x.max \leftarrow x $\;
        $MaxMinCoord[P_n^{(t,j)}].x.min \leftarrow x $\;
        % $MaxMinCoord[net_{id}].y.max \leftarrow y $\;
        % $MaxMinCoord[net_{id}].y.min \leftarrow y $\;
    } {
        \tcp{\textcolor{blue}{Update the bounding box range}}
        \uIf{$MaxMinCoord[P_n^{(t,j)}].x.max < x $} {
        $reward \leftarrow reward +(x-MaxMinCoord[P_n^{(t,j)}].x.max)$\;
        $MaxMinCoord[P_n^{(t,j)}].x.max = x$\;}
        \ElseIf{$MaxMinCoord[P_n^{(t,j)}].x.min > x $}{
        $reward \leftarrow reward + (MaxMinCoord[P_n^{(t,j)}].x.min - x)$ \;
        $MaxMinCoord[P_n^{(t,j)}].x.min = x$\;
        }
    }
}
\end{algorithm}

\paragraph{Position Mask Generation.}

The efficient position mask generation algorithm is in Algorithm \ref{algorithm2}.
% \subsection{Position Mask Generation Algorithm}
\begin{algorithm}
\caption{Position Mask Generation}
\label{algorithm2}
\KwData{Width, Height and Position of t-1 placed module $M^{1:t-1}$ $(M_w^{1:t-1}, M_h^{1:t-1}, M_x^{1:t-1}, M_y^{1:t-1})$}
\KwResult{Position Mask $f^t_p$ for Module $M^t$}
$f^t_p\leftarrow ones(N,N)$\tcp*[l]{\textcolor{blue}{$ones(N,N)$ is all-ones $N\times N$ matrix}}\
\For{$i\leftarrow 1$ \KwTo $t-1$}{
    $tmp \leftarrow ones(N, N)$\;
    \tcp{\textcolor{blue}{find positions that  will cause $M^t$ and $M^i$ to overlap}}
    $tmp[M^i_x - M^t_w +1: M^i_x + M^i_w -1, M^i_y - M^t_h +1: M^i_y + M^i_h -1]\leftarrow 0$\;
    \tcp{\textcolor{blue}{exclude infeasible positions}}
    $f^t_p \leftarrow tmp \odot f^t_p$ \tcp*[l]{\textcolor{blue}{$\odot$ is element-wise product}}\
}
% $f^t_p \leftarrow 1-f^t_p$
\end{algorithm}

\paragraph{Wire Mask Generation.}

The efficient wire mask generation algorithm is shown in Algorithm \ref{algorithm3}. For simplicity, we omit the calculation of the y dimension, which is the same as the x dimension.

% \subsection{Wirelength Mask Generation Algorithm}
\begin{algorithm}
\caption{Wire Mask Generation (omit y-dimension)}
\label{algorithm3}
\KwData{Hash Map of Max/Min X/Y coordinates of nets $MaxMinCoord$,
pin's offsets $(\Delta_x^{(t,j)}, \Delta_y^{(t,j)})$, pin to net connection $P_n^{(t,j)}$}
\KwResult{Wire Mask $f_w^t$ for module $M^t$}
$f_w^t\leftarrow zeros(N,N)$\;
\tcp{\textcolor{blue}{Accumulate the wirelength increase for each net}}
\ForEach{$\Delta_x^{(t,j)}, P_n^{(t,j)}$ of all pins $P^{(t,j)}$ from $M^{t}$ }{
    \tcp{\textcolor{blue}{If the pin is to the left of the net bounding box}}
    \For{$i \leftarrow 0$ \KwTo $MaxMinCoord[P_n^{(t,j)}].x.min + \Delta_x^{(t,j)} -1$}{
        $f_w^t[i, :] \leftarrow f_w^t[i, :] + MaxMinCoord[P_n^{(t,j)}].x.min + \Delta_x^{(t,j)} - i$\;
    }
    \tcp{\textcolor{blue}{If the pin is to the right of the net bounding box}}
    \For{$i \leftarrow MaxMinCoord[P_n^{(t,j)}].x.max + \Delta_x^{(t,j)} + 1$ \KwTo $N-1$}{
        $f_w^t[i, :] \leftarrow f_w^t[i, :] + i - (MaxMinCoord[P_n^{(t,j)}].x.max + \Delta_x^{(t,j)})$\;
    }
}
\end{algorithm}

\paragraph{Congestion Satisfaction.}

The algorithm implemented in the congestion satisfaction block can be seen in Algorithm \ref{algorithm4}.
% \subsection{Congestion Search}
\label{congestionsearch}
\begin{algorithm}
\caption{Placement with Congestion Constraint}\label{algorithm4}
\KwData{Trained place agent $agent$, expected congestion threshold $C_{th}$}
\KwResult{A placement plan $[a_1, a_2, ..., a_{V}]$ that meet the congestion requirement}
% Initialize $state$\;
\For{$i \leftarrow 1$ \KwTo $V$}{
Choose $a_i$ from the probability matrix generated by policy network $agent$\;
% $state_{tmp} \leftarrow $ observe state when taking $action_{i}$\;
$Cong \leftarrow$ congestion matrix from the state after taking $a_i$\;
Compute congestion value $c$ from $Cong$\;
\If{$c > C_{th}$}{
    Randomly sample $N$ different actions $a_i^{1:N}$ from action space\;
    Compute $N$ congestion values $c_i^{1:N}$ from congestion metrics\;
    Get $N$ wirelength values $w_i^{1:N}$ from wire masks\;
    Sort the $N$ actions according to $w_i^{1:N}$ (as the 1st key) and $c_i^{1:N}$ (as the 2nd key)\;
    $flag \leftarrow False$ \;
    \For{$j \leftarrow 1$ \KwTo $N$}{
        \If {$c_i^j \le C_{th}$}{
            $flag \leftarrow True$\;
            $a_i \leftarrow a_i^j$\;
            break\;
        }
    }
    \tcp{\textcolor{blue}{If all sampled actions cannot satisfy congestion threshold, we choose the one with minimal congestion increase.}}
    \lIf{$ flag$ is $False$}{ $a_i \leftarrow $ the action $a_i^j$ with minimum $c_i^j$}
}
    Take action $a_i$ as the final action\;
    % Update $state$ \; 
}
\end{algorithm}

\subsection{Details of Model Architecture}
\label{detailarch}
The parameters of layers in model architecture are in Table \ref{aarchi}. Also, the features used by pixel-level mask generation are in Table \ref{state}. The comparison of features for the placement order in different methods can be seen in Table \ref{tab:order}.

\begin{table}[!ht]
	\caption{Model Architecture}
	\label{aarchi}
    \centering
    \begin{tabular}{cccc}
    \toprule
        Block & Layer & Kernel Size & Output shape \\ \midrule
        \multirow{3}{*}{Local Mask Fusion}&Conv&$1\times 1$& $(224, 224, 8)$\\
        &Conv&$1\times 1$& $(224, 224, 8)$\\
        &Conv&$1\times 1$& $(224, 224, 1)$\\\midrule
        \multirow{2}{*}{Global Mask Encoder}&ResNet-18&-& 1000\\
        &FC&-& 768\\ \midrule
        \multirow{5}{*}{Global Mask Decoder}&Deconv& $3\times3$& $(14, 14, 8)$\\
        &Deconv& $3\times3$& $(28, 28, 4)$\\
        &Deconv& $3\times3$& $(56, 56, 2)$\\
        &Deconv& $3\times3$& $(112, 112, 1)$\\
        &Deconv& $3\times3$& $(224, 224, 1)$\\ \midrule
        Merge & Conv & $1\times1$& $(224, 224, 1)$\\ \midrule
        Position Embedding & - &  - & $64$ \\ \midrule
        \multirow{3}{*}{FC for Value} & FC & - &$512$\\
         & FC & - &$64$\\
          & FC & - &$1$\\
        \bottomrule
    \end{tabular}
\end{table}

\begin{table}[!ht]
	\caption{State Features}
	\label{state}
    \centering
    \begin{tabular}{ccccc}
    \toprule
        Module status & Index & Feature & Notation & Dimension per module\\ \midrule
        \multirow{5}{*}{Placed}&  \multirow{5}{*}{$M^{1:t-1}$} & Width & $M_w$ & 1\\
        & & Height & $M_h$ & 1\\
        & & Position & $M_x, M_y$ & 2\\
        & & Pin Offset & $\Delta_x, \Delta_y$ & $2\times \mbox{num of pins}$ \\
        & & Pin to Net Connection & $P_n$ & $\mbox{num of pins}$ \\\midrule
        \multirow{4}{*}{Unplaced} & \multirow{4}{*}{$M^t, M^{t+1}$} & Width & $M_w$ & 1\\
        & & Height & $M_h$ & 1\\
        & & Pin Offset & $\Delta_x, \Delta_y$ & $2\times \mbox{num of pins}$ \\
        & & Pin to Net Connection & $P_n$ & $\mbox{num of pins}$ \\
        \bottomrule
        
    \end{tabular}
\end{table}

\begin{table}[!ht]
\centering
\caption{Features used for placement order}
\small
\label{tab:order}
\begin{tabular}{cc}\\\toprule
Method & Features for place order \\\midrule
Graph Placement \cite{mirhoseini2021graph} & Topological order, Area \\  
DeepPR \cite{cheng2021joint} & None \\  
MaskPlace & Number of nets, Area, Number of its connected modules have been placed\\  
\bottomrule
\end{tabular}
\end{table} 

\subsection{Training Configuration}
\label{detailtrain}
The detailed configuration and hyperparameter settings of our model is in Table \ref{aconfig}.
\begin{table}[!ht]
	\caption{Model Configuration}
	\label{aconfig}
    \centering
    \begin{tabular}{cccc}
    \toprule
        Configuration & Value &Configuration & Value\\ \midrule
        Optimizer & Adam & Learning rate & $2.5\times 10^{-3}$\\
        Total epoch & 150 &  Epoch for update & 10\\ 
        Batch size & 64 & Buffer capacity & $10\times \mbox{num of modules}$\\
        Clip $\epsilon$ & 0.2 & Clip gradient norm & 0.5 \\
        Reward discount $\gamma$ &0.95& Num GPUs & 1\\
        CPU & AMD Ryzen 9 5950X & GPU & GeForce RTX 3090 \\
        \bottomrule
    \end{tabular}
\end{table}

Also, we implement DREAMPlace\footnote{\href{https://github.com/limbo018/DREAMPlace}{github.com/limbo018/DREAMPlace}} \cite{lin2020dreamplace}, Graph Placement\footnote{\href{https://github.com/google-research/circuit_training}{github.com/google-research/circuit\_training}} \cite{mirhoseini2021graph} ,and DeepPR\footnote{\href{https://github.com/Thinklab-SJTU/EDA-AI}{github.com/Thinklab-SJTU/EDA-AI}} \cite{cheng2021joint} by their open source codes with their default settings.

\subsection{Details of Benchmark}
\label{morebench}
The detailed statistics of benchmarks are in Table \ref{benchmark}.
Hard macros are macros placed by the RL method in Graph Placement \cite{mirhoseini2021graph}, and the remaining macros, also named soft macros, are placed by the classic optimization-based method. However, this distinction does not apply to the process of our method, which means we place all macros by RL.
The statistical range of nets, pins, and area utilization are macros. Ports are terminals connecting to an external circuit, seen as fixed and no-size modules. Our method is also applicable to circuits with ports without additional modifications.

\begin{table}[!ht]
 \small
	\caption{Statistics of different chip benchmarks.}
	\label{benchmark}
	\centering
	\begin{tabular}{cccccccc}
	
		\toprule
% 		\multicolumn{2}{c}{Part}                   \\
% 		\cmidrule(r){1-2}
		Benchmark & Macros & Hard Macros & Standard Cells & Nets & Pins & Ports & Area Util(\%)  \\
		\midrule
		adaptec1 & 543 & 63 & 210,904 & 3,709 & 4,768 & 0 & 55.62  \\ 
        adaptec2 & 566 & 190 & 254,457 & 4,346 & 10,663 & 0 & 74.46  \\ 
        adaptec3 & 723 & 201 & 450,927 & 6,252 & 11,521 & 0 & 61.51  \\ 
        adaptec4 & 1,329 & 92 & 494,716 & 5,939 & 13,720 & 0 & 48.62  \\ 
        bigblue1 & 560 & 32 & 277,604 & 657 & 1,897 & 0 & 31.58  \\ 
        bigblue3 & 1,293 & 138 & 1,095,519 & 5,537 & 15,225 & 0 & 66.81  \\ 
        ariane & 932 & 134 & 0 &12,404 & 22,802 & 1,231 & 78.39  \\ 
        ibm01 & 246 & 246 & 12,506 & 908 & 1,928 & 246 & 61.94 \\ 
        ibm02 & 280 & 272 & 19,321 & 602 & 1,466 & 259 & 64.63 \\ 
        ibm03 & 290 & 290 & 22,846 & 614 & 1,237 & 283 & 57.97 \\ 
        ibm04 & 608 & 296 & 26,899 & 1,512 & 3,167 & 287 & 54.88 \\ 
        ibm06 & 178 & 178 & 32,320 & 83 & 175 & 166 & 54.77 \\ 
        ibm07 & 507 & 292 & 45,419 & 2,471 & 5,992 & 287 & 46.03 \\ 
        ibm08 & 309 & 302 & 51,000 & 1,725 & 3,721 & 286 & 47.13 \\ 
        ibm09 & 253 & 56 & 53,142 & 446 & 898 & 285 & 44.52 \\ 
        ibm10 & 786 & 56 & 68,643 & 2,160 & 4,720 & 744 & 61.40 \\ 
        ibm11 & 373 & 56 & 70,185 & 682 & 1,371 & 406 & 41.40 \\ 
        ibm12 & 651 & 205 & 70,425 & 1,589 & 3,468 & 637 & 53.85 \\ 
        ibm13 & 424 & 100 & 83,775 & 804 & 1,669 & 490 & 39.43 \\ 
        ibm14 & 614 & 91 & 146,991 & 1,620 & 3,960 & 517 & 22.49 \\ 
        ibm15 & 393 & 22 & 161,177 & 748 & 1,521 & 383 & 28.89 \\ 
        ibm16 & 458 & 37 & 183,026 & 1,755 & 3,981 & 504 & 39.46 \\ 
        ibm17 & 760 & 107 & 184,735 & 2,055 & 4,366 & 743 & 19.11 \\ 
        ibm18 & 285 & 285 & 210,328 & 727 & 1,600 & 272 & 11.09 \\ 
		\bottomrule
	\end{tabular}
% 	}
	
\end{table}

\subsection{Supplementary Experiment}
\label{suppexp}

\paragraph{More benchmarks}
We also conducted experiments in the IBM benchmark suite (ICCAD 2004) \cite{adya2009iccad}, which has been used to evaluate placement for more than a decade.
% This benchmark suite comprises 18 chip designs with 178\textasciitilde786 macros and 12k\textasciitilde210k standard cells. 
We remove the ``ibm05'' because it does not contain any macros. We use our MaskPlace to place large macros and DREAMPlace \cite{lin2020dreamplace} to place standard cells. We compared our method with graph placement \cite{mirhoseini2021graph} and the simulated annealing method used in \cite{mirhoseini2021graph}. The results are in Table \ref{ibm} and our method can achieve the lowest HPWL in all benchmarks.

% For the \textbf{} 

\begin{table}[!ht]
	\caption{Comparisons of HPWL ($\times 10^5$) for macro and standard cell placement in ibm benchmark.}
	\label{ibm}
    \centering
    \begin{tabular}{ccccccc}
    \toprule
        Method & ibm01 & ibm02 & ibm03 & ibm04 & ibm05 & ibm06   \\ \midrule
        Graph Placement \cite{mirhoseini2021graph} & 31.71 & 55.12 & 80.00 & 86.86 & - & 63.48  \\ 
        Simulated Annealing
 \cite{mirhoseini2021graph} & 25.85 & 54.87 & 80.68 & 83.32 & - & 69.09  \\ 
        MaskPlace+DREAMPlace \cite{lin2020dreamplace} & \textbf{24.18} & \textbf{47.45} & \textbf{71.37} & \textbf{78.76} & - & \textbf{55.70}  \\  
        \toprule
        Method & ibm07 & ibm08 & ibm09 & ibm10 & ibm11 & ibm12  \\ \midrule
        Graph Placement \cite{mirhoseini2021graph} & 117.71 & 134.77 & 148.74 & 440.78 & 218.73 & 438.57  \\ 
        Simulated Annealing
 \cite{mirhoseini2021graph} & 117.71 & 144.89 & 141.67 & 463.04 & 228.79 & 435.77  \\ 
        MaskPlace+DREAMPlace \cite{lin2020dreamplace} & \textbf{95.27} & \textbf{120.64} & \textbf{122.91} & \textbf{367.55} & \textbf{202.23} & \textbf{397.25}  \\ 
        \toprule
        Method & ibm13 & ibm14 & ibm15 & ibm16 & ibm17 & ibm18  \\ \midrule
        Graph Placement \cite{mirhoseini2021graph} & 278.93 & 455.31 & 520.06 & 642.08 & 814.37 & 450.67  \\ 
        Simulated Annealing
 \cite{mirhoseini2021graph} & 259.89 & 405.80 & 510.06 & 614.54 & 720.40 & 442.00  \\ 
        MaskPlace+DREAMPlace \cite{lin2020dreamplace} & \textbf{246.49} & \textbf{302.67} & \textbf{457.86} & \textbf{584.67} & \textbf{643.75} & \textbf{398.83}  \\ \bottomrule 
    \end{tabular}
\end{table}

For the larger circuit \textit{bigblue4} in ISPD 2005 benchmark, the result of our method and baselines can been seen as Table \ref{bigblue4}. MaskPlace still achieved the best performance.

\begin{table}[!ht]
	\caption{\small{HPWL ($\times 10^7$) results for \textit{bigblue4} benchmark}}
	\label{bigblue4}
    \centering
    \small
    % \resizebox{\textwidth}{13mm}{
    % \begin{adjustbox}{max width=0.98\textwidth}
    \begin{threeparttable}
    {
    \begin{tabular}{ccccc}
    
    \toprule
        Benchmark  & Random & NTUPlace3\cite{chen2008ntuplace3} &RePlAce\cite{cheng2018replace} &  DREAMPlace \cite{lin2020dreamplace} \\ \midrule
        bigblue4 & 128.06±3.94 & 48.38 & 11.80±0.73 & 12.29±1.64 \\ \midrule
        Benchmark & Graph Placement \cite{mirhoseini2021graph} & DeepPR \cite{cheng2021joint}& DeepPR-no-overlap \cite{cheng2018replace}& MaskPlace \\ \midrule
        bigblue4 & 53.35±4.06 & 68.30±4.44 & 115.08±2.29 & \textbf{11.07±0.90} \\
      
        % \midrule
        % DeepPR  (Routing) & ~ & ~ & ~ & ~ & ~ &   \\ 
        % MaskPlace (Routing) & ~ & ~ & ~ & ~ & ~ &   \\ 
        \bottomrule
    \end{tabular}
    }
    % \begin{tablenotes}
    % % \item[*] We modify the raw DeepPR method to take real size of macros into consideration and avoid overlaps. 
    % % \small
    % % \item[*] Compared with the HPWL result from the model trained on the corresponding benchmark.
    % \end{tablenotes}
    \end{threeparttable}
    % \end{adjustbox}
    % \caption{HPWL results of methods}
    % }
\end{table}

\paragraph{Search time}

\begin{figure}[!ht]
\centering
% \pluo{rewrite caption make it self-contain, expalin and discuss more.}
\includegraphics[width=0.5\textwidth]{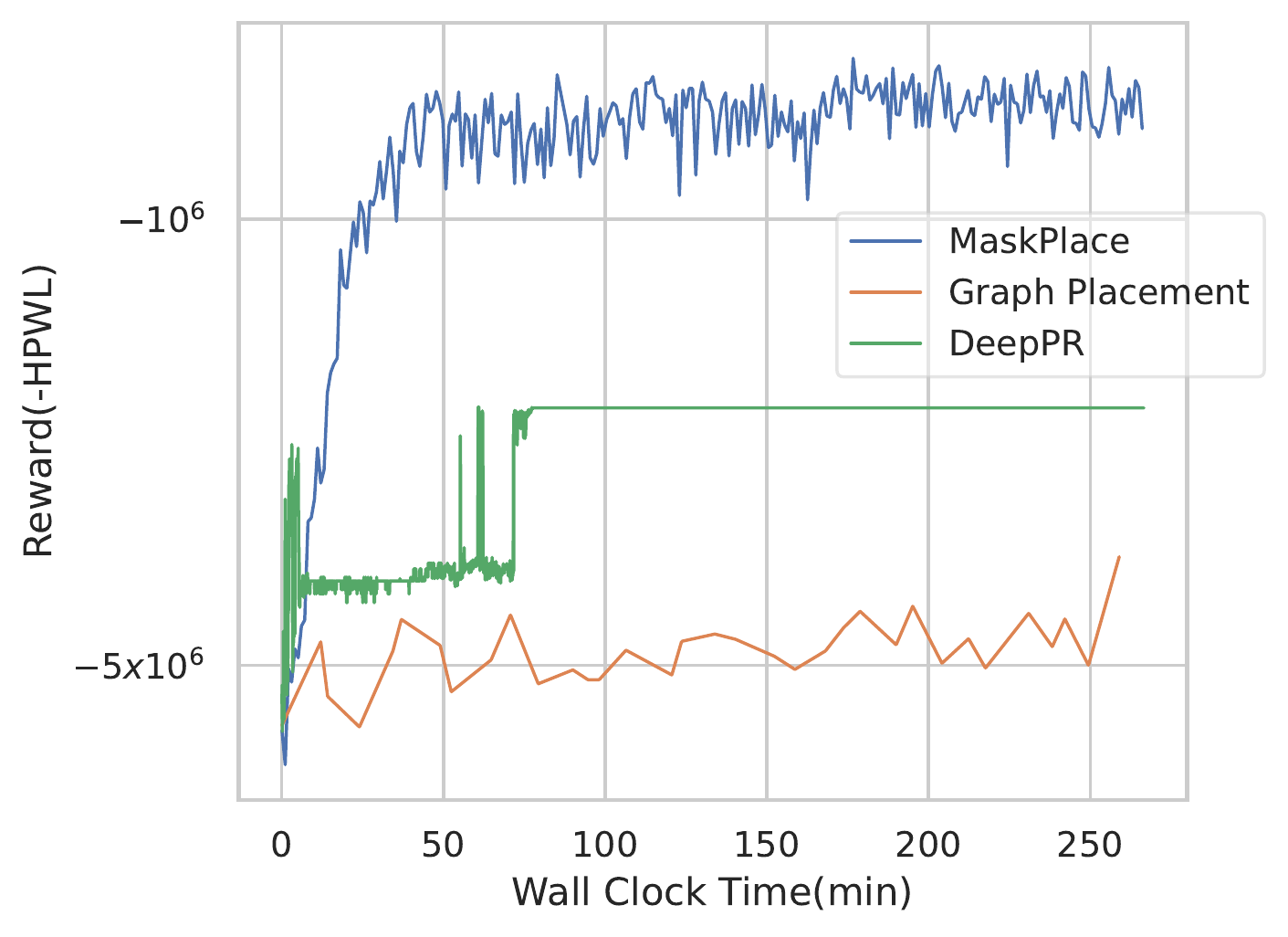}
\caption{Search time comparison}
%Fusion is the information exchange between position mask and wirelength mask. \textcolor{red}{$\triangle$} is the position with high probability to place (no overlap and shorter wirelength).

% \vspace{-8pt}
\label{clocktime}
\end{figure}

We compared the search time of our method, Graph Placement \cite{mirhoseini2021graph} and DeepPR \cite{cheng2021joint}. We tested all methods in the same environment and took the HPWL as the metric in benchmark \textit{adaptec1}. The result is in Fig. \ref{clocktime}. We can see that our approach can achieve the best performance in a few hours.

\subsection{Detailed equation description of the model}
We describe our model architecture in Fig. \ref{architecture} in the form of equation.

With current state $s_t$, we first calculate the position masks $f_p^t,f_p^{t+1}$, wire masks $f_w^t,f_w^{t+1}$ and view mask $f_v^t$ via the mask generation function $m(\cdot)$.
\begin{equation}
f_p^t,f_p^{t+1},f_w^t,f_w^{t+1},f_v^t=m(s_t)
\end{equation}
Then we extract the local feature $z_t^l$ via local mask fusion $g_\omega(\cdot)$ and the global features $z_t^g$ via the global mask encoder $enc_\eta(\cdot)$.
\begin{equation}
z_t^l=g_\omega(f_p^t,f_p^{t+1},f_w^t,f_w^{t+1})
\end{equation}
where $g_\omega(\cdot)$ is a 1×1 convolutional neural network with parameter $\omega$. 
\begin{equation}
z_t^g=enc_\eta(f_w^t,f_w^{t+1},f_v^t)
\end{equation}
where $enc_\eta(\cdot)$ is a convolutional neural network with ResNet-18 architecture with parameter $\eta$.

With local features $z_t^l$ and global features $z_t^g$, the state value $\hat{V}_t$ is derived by
\begin{equation}
\hat{V}_t=v_\phi(pos(t), z_t^g)
\end{equation}
where $v_\phi$ is an MLP-like neural network with parameter $\phi$ and $pos(t)$ is an embedding vector which is related to step $t$.

We decode the global features $z_t^g$ into the dimension as same as the action space by the global mask decoder
\begin{equation}
z_t^{'g}=dec_\delta(z_t^g)
\end{equation}
where $dec_\delta(\cdot)$ is a transpose convolutional neural network with parameter $\delta$.

Finally, we concatenate the local features $z_t^l$ and global features $z_t^{'g}$ in the channel dimension and merge them by another 1×1 convolutional neural network $\psi_\xi(\cdot)$. We further combine it with the position mask $f_p^t$ to generate action $a_t$ via the policy network $\pi_\theta(\cdot)$
\begin{equation}
a_t\sim\pi_\theta(\psi_\xi(z_t^l||z_t^{'g}),f_p^t)
\end{equation}
where $\pi_\theta(\cdot)$ is an MLP-like neural network with parameter $\theta$ and $\psi_\xi(\cdot)$ is a 1×1 convolutional neural network with parameter $\xi$.

% \subsection{Features for placement order comparison}

% Different features chosen for placement order can be seen in Tab. \ref{tab:order}. Our method introduces the number of nets feature, which can improve the quality of placement.

\section{Related Work}
\label{Related Work}
\paragraph{Classic optimization-based methods.} Optimization has been the dominant method in placement for decades. They can be divide into three categories: partitioning-based methods \cite{roy2006min, khatkhate2004recursive}, simulated annealing methods \cite{yang2000dragon2000, vashisht2020placement} and analytical methods \cite{chen2008ntuplace3, lu2014eplace, cheng2018replace, lin2020dreamplace, viswanathan2007rql, viswanathan2007fastplace, kim2012maple, kim2012complx, brenner2015bonnplace, lin2013polar, spindler2008kraftwerk2, chan2006mpl6, kahng2005aplace, gu2020dreamplace}. 

Partitioning-based methods \cite{roy2006min, khatkhate2004recursive} cluster the whole circuits into several parts to minimize the connections between parts. These methods first solve the placement problems within the same part and then place these parts to suitable positions on the chip based on the divide-and-conquer idea. However, optimizing modules within one part is an isolated problem, and sometimes it is hard to divide the circuits into relatively independent parts, which is highly related to the topology of the circuits.

Simulated Annealing (SA) methods \cite{yang2000dragon2000, vashisht2020placement} are also known as hill-climbing methods, a widely used iterative heuristic algorithm for solving combinatorial optimization problems. They initialize a random status and then search for the following status by moving from the current status to a neighbor status. If the metrics of the neighbor status are better than that of the current status, they move to the neighbor status. Otherwise, the move may still be taken with a decreasing probability over time. The advantage is that they can be implemented when metrics do not have the analysis formula or cannot be differentiable. However, it is not efficient enough, and the placement results are highly dependent on the random initial state.

Analytical methods gradually replace the above two methods because of the best performance. They can be divided into quadratic methods \cite{viswanathan2007rql, viswanathan2007fastplace, kim2012maple, kim2012complx, brenner2015bonnplace, lin2013polar, spindler2008kraftwerk2} and nonlinear (non-quadratic) methods \cite{chen2008ntuplace3, lu2014eplace, cheng2018replace, lin2020dreamplace, chan2006mpl6, kahng2005aplace,  gu2020dreamplace}. 
Quadratic methods \cite{viswanathan2007rql, viswanathan2007fastplace, kim2012maple, kim2012complx, brenner2015bonnplace, lin2013polar, spindler2008kraftwerk2} transform the placement problem into a sequence of convex quadratic problems, and there are well-established solvers for such problems. However, it is a very rough approximation.
Nonlinear methods \cite{chen2008ntuplace3, lu2014eplace, cheng2018replace, lin2020dreamplace, chan2006mpl6, kahng2005aplace,  gu2020dreamplace} design a single differentiable objective function and optimize it. The advantage is that it can handle large-scale modules. However, the objective function is still approximated, and they cannot avoid overlaps when combining multiple metrics in one objective function. Methods in this category achieved the highest placement quality among all classic methods \cite{lin2020dreamplace}.

\paragraph{Learning-based methods.}
With the development of deep learning, some learning-based approaches \cite{vashisht2020placement, huang2019routability, kirby2021guiding,  agnesina2020vlsi} have been proposed to assist classic methods. \citet{huang2019routability} uses convolutional neural networks to estimate the congestion for SA placement. \citet{vashisht2020placement} uses the reinforcement learning models to generate the initial placement of SA. \citet{kirby2021guiding, agnesina2020vlsi} help classic placement tools choose the most suitable hyperparameters with reinforcement learning methods. However, these methods do not implement end-to-end placement by deep learning, so the placement results depend heavily on the classic methods.

Pure reinforcement learning methods \cite{mirhoseini2021graph, cheng2021joint, jiang2021delving, chang2022flexible} view placement as a process of placing modules sequentially. \citet{mirhoseini2021graph} uses reinforcement learning to place hard macros, and the force-directed method \cite{spindler2008kraftwerk2} to place remaining soft macros. \citet{jiang2021delving} replaces the force-directed method with DREAMPlace \cite{lin2020dreamplace} to place soft macros based on Graph Placement \cite{mirhoseini2021graph}. \citet{cheng2021joint} proposes a reinforcement learning method by using wirelength as the reward. Moreover, \citet{chang2022flexible} puts all metrics in the RL reward. They have in common that they convert the circuit as a graph structure and input them to the graph neural networks \cite{kipf2016semi}. However, the pin information has been lost, leading to sub-optimal placement. Also, they cannot avoid overlaps because of the reduction in search space. These methods still have room for improvement in terms of realistic chip placement. For instance, DeepPR \cite{cheng2021joint} ignores the realistic size of the module. However, the size of the modules varies widely in most circuits. Although it proposes to use routing wirelength instead of HPWL as the reward, it will affect the efficiency and lead to sparse reward, making models hard to train. In contrast, HPWL is a high-quality wirelength estimation, and we do not need to discard this inherent dense reward.

% There are placement methods with the help of deep learning \cite{huang2019routability, kirby2021guiding}, but most of them use AI technology to assist traditional methods.
\end{document}